\documentclass[twoside]{article}

\usepackage[accepted]{aistats2024}
\usepackage[round]{natbib}

\usepackage{amsmath}
\usepackage{amssymb}
\usepackage{amsfonts}
\usepackage{amstext}
\usepackage{amsthm}
\usepackage{algorithm}
\usepackage{algorithmicx}
\usepackage{algpseudocode}
\usepackage{xcolor}
\usepackage{bbm}
\usepackage{mathtools}
\usepackage{multirow}
\usepackage{diagbox}
\usepackage{url}
\newcommand{\bx}{\mathbf{x}}

\newcommand{\msafeucb}{\textsc{\sffamily M-SafeUCB }}
\newcommand{\msafeopt}{\textsc{\sffamily M-SafeOpt }}
\newcommand{\msafeoptns}{\textsc{\sffamily M-SafeOpt}}

\newcommand{\predvar}{\textsc{\sffamily PredVar }}
\newcommand{\predvarns}{\textsc{\sffamily PredVar}}
\newcommand{\safeopt}{\textsc{\sffamily SafeOpt }}
\newcommand{\safeoptns}{\textsc{\sffamily SafeOpt}}
\newcommand{\safeoptmc}{\textsc{\sffamily SafeOpt-MC }}
\newcommand{\safeoptmcns}{\textsc{\sffamily SafeOpt-MC}}
\newcommand{\stageopt}{\textsc{\sffamily StageOpt }}
\newcommand{\stageoptns}{\textsc{\sffamily StageOpt}}
\newcommand{\goose}{\textsc{\sffamily GOOSE }}
\newcommand{\dDX}{\mathcal{D_X}}
\newcommand{\dDS}{\mathcal{D_S}}
\newcommand{\dD}{\mathcal{D}}
\newcommand{\UCB}{\mathrm{UCB}}

\newcommand{\LCB}{\mathrm{LCB}}
\newcommand{\GP}{\mathrm{GP}}
\newcommand{\usxt}{\underline{s}^{(\bx)}_t}

\DeclareMathOperator*{\argmax}{arg\,max}

\newtheorem{theorem}{Theorem}

\newtheorem{lemma}{Lemma}

\begin{document}

\runningtitle{No-Regret Algorithms for Safe Bayesian Optimization}

%

%

\twocolumn[

\aistatstitle{No-Regret Algorithms for Safe Bayesian\\  Optimization with Monotonicity Constraints}

\aistatsauthor{ Arpan Losalka \And Jonathan Scarlett }

\aistatsaddress{ National University of Singapore \And  National University of Singapore } ]

\begin{abstract}
    We consider the problem of sequentially maximizing an unknown function $f$ over a set of actions of the form $(s,\bx)$, where the selected actions must satisfy a safety constraint with respect to an unknown safety function $g$.  We model $f$ and $g$ as lying in a reproducing kernel Hilbert space (RKHS), which facilitates the use of Gaussian process methods.  While existing works for this setting have provided algorithms that are guaranteed to identify a near-optimal safe action, the problem of attaining low cumulative regret has remained largely unexplored, with a key challenge being that expanding the safe region can incur high regret.  To address this challenge, we show that if $g$ is monotone with respect to just the \textit{single variable} $s$ (with no such constraint on $f$), sublinear regret becomes achievable with our proposed algorithm.  In addition, we show that a modified version of our algorithm is able to attain sublinear regret (for suitably defined notions of regret) for the task of finding a near-optimal $s$ corresponding to every $\bx$, as opposed to only finding the global safe optimum.  Our findings are supported with empirical evaluations on various objective and safety functions.
\end{abstract}

\section{INTRODUCTION}

Sequential optimization of an unknown function is an important task with many applications, and comes with an interesting set of challenges in the scenario that function queries are expensive. Bayesian optimization is a popular approach for this task, with applications including robotics \citep{lizotte2007automatic}, environmental monitoring \citep{srinivas2012information}, adaptive clinical trial design \citep{takahashi2021mtd}, hyperparameter tuning in machine learning \citep{snoek2012practical} and recommendation systems \citep{vanchinathan2014explore}, among others.

This black-box optimization problem becomes even more challenging when considering a safety constraint along with the optimization objective. One important notion of safety that has been considered in the literature is to only allow actions for which an unknown safety function takes values above a pre-defined safety threshold \citep{sui2015safe}. Various methods have been proposed for this task, such as \safeopt \citep{sui2015safe}, \stageopt \citep{sui2018stagewise} and \safeoptmc \citep{berkenkamp2021bayesian}, among others. The core idea of these algorithms is to start from a safe seed set of inputs, and cautiously expand the set and identify the most promising regions within it, in order to reach the optimal action.  Accordingly, the main performance metrics considered have been \emph{expanding the safe set as much as possible} and \emph{returning a single near-optimal action}.  In contrast, the goal of small \emph{cumulative regret} has remained relatively unexplored in safe settings, despite being the most widely-adopted performance measure in the vanilla setting.  A key difficulty in the safe setting is that expanding the set of known safe points is ``purely explorative'' and may require sampling many highly suboptimal points.

In this work, we consider the scenario where the safety function $g$ is known to increase monotonically with respect to a safety variable $s \in \dDS$, while the objective function $f$ is distinct from $g$ and need not have any such structure.  We show that with this mild assumption, strong guarantees on the cumulative regret become possible.  We also introduce other notions of regret associated with finding the best $s$ for \textit{every $\bx \in \dDX$ separately}, and we show that our algorithms can be simplified (while maintaining similar guarantees) when \emph{both} $f$ and $g$ are monotone in $s$.

\paragraph{Motivating Applications:} 
Consider the problem of an adaptive Phase I/II clinical trial design for the purpose of finding drug doses that simultaneously satisfy safety constraints with respect to drug toxicity, as well as achieve optimal efficacy \citep{berry2012adaptive}. A common structure exhibited by the toxicity of various classes of drugs is that it increases monotonically as a function of the drug dosage \citep{chevret2006statistical}. While the efficacy may also increase similarly, it is not the case in general, especially when drug combinations are used \citep{cai2014bayesian}. In such a scenario, the problem of finding the optimal safe drug dose fits our problem formulation, since both toxicity ($g$) and efficacy ($f$) are unknown functions of the drug doses (with $s$ denoting dosage of one drug, and $\bx$ denoting that of other drugs). Here, we need to optimize $f$ while satisfying the toxicity threshold $g(s,\bx) \leq h$, and regret minimization in this setting implies that we not only find the optimum (minimizing simple regret), but we also maximize benefits to the trial participants simultaneously (minimizing cumulative regret). 

The requirement of a trial may also be to find the optimal dose for a range of patient's characteristics (e.g., age group, gender, etc.). In this case, our alternate problem setting is relevant, where we want to find the optimum $s$ (dose) for every $\bx$ (patient's characteristics), while satisfying safety constraints (toxicity threshold). The problem formulation also extends naturally when considering drug-combinations, as long as monotonic behavior of drug toxicity holds with respect to \textit{at least one drug dose}. 

Another application area suited to the goal of choosing parameters to optimize performance (with respect to $f$) while ensuring safety (with respect to $g$) is robotics \citep{berkenkamp2017safe}.  Here, the notion of a ``safety variable'' $s$ might be explicitly incorporated into the system by design, i.e., we have a controllable variable that directly dictates ``how cautiously'' the task is performed.  Alternatively, certain variables such as acceleration and torque might be \emph{implicitly} connected to safety and naturally provide our required monotonicity constraint.

\paragraph{Related Work:} The problem of safe Bayesian optimization was first considered by \cite{sui2015safe}, who proposed the \safeopt algorithm for this task. This algorithm, as well as other algorithms that were proposed subsequently such as \stageopt \citep{sui2018stagewise}, \safeoptmc \citep{berkenkamp2021bayesian}, and \goose \citep{turchetta2019safe}, aims to expand a safe seed set widely enough to guarantee identifying a near-optimal safe point (excluding those from regions that are ``unreachable'').  
Extensions have also been provided to reinforcement learning \citep{turchetta2016safe,berkenkamp2017safe,turchetta2020safe}. 

A distinct approach that can give low cumulative regret was proposed in \citep{amani2021regret}, but since it expands the safe set in a ``one-shot'' manner, it is only suited to kernels that are finite-dimensional or extremely smooth.   Improvements over the above algorithms are also possible for safety functions modeled by dynamical systems \citep{Baumann2021GoSafeGO,Sukhija2022ScalableSE}, but our focus is on static functions.  We refer the reader to \citep{losalka2023benefits} for further discussion on all of these works.

The closest work to ours is that of \citet{losalka2023benefits}, who also use monotonicity of the unknown function $f$ with respect to a safety variable $s$, and design the \msafeucb algorithm. While they prove a sublinear regret bound, the applicability of the algorithm is limited by the fact that the safety constraint is defined with respect to $f$ itself.  In the more general setting where $f$ and $g$ are distinct, their algorithm is only directly suitable when \emph{both $f$ and $g$} are monotone in $s$ (see Section \ref{sec:algo} for further details). A naive strategy for overcoming this would be to explore using $g$ and then pass the resulting safe set to an optimizer for $f$, but the former step may already incur high cumulative regret with respect to $f$.  Overall, our more general setting significantly complicates both the algorithm design and the theoretical analysis. 

\paragraph{Contributions:} Our main contributions are summarized as follows.
\begin{enumerate} \itemsep0ex
    \item We introduce the problem of safe optimization of an unknown function $f$ when the safety function $g$ is known to be monotone with respect to a safety variable $s$. We propose an algorithm, \msafeoptns, which uses a novel acquisition function designed to balance exploration and exploitation, along with the elimination of provably suboptimal $\bx$'s, to attain sublinear cumulative regret.
    \item We consider an alternative setting where the goal is to find the optimal action corresponding to \emph{every} $\bx \in \dDX$ (i.e., the best safe $s$ for every $\bx$), and adapt our algorithm for this task.  We provide modified cumulative regret notions capturing both the degree of optimality of sampled points and a \emph{worst-case} notion over all $\bx$ (see Section \ref{sec:problem} for details), with both guaranteed to be sublinear.
    \item We show how our algorithm can be simplified in the scenario that both $f$ and $g$ increase monotonically with respect to $s$, while attaining the same theoretical guarantees for the goal of safe optimization across the input domain. Alternatively, when the goal is to optimize $s$ for every $\bx$, we show how further simplification recovers the \msafeucb algorithm and its guarantees.
    \item We empirically evaluate our proposed algorithms on various synthetic functions, showing significant improvements over several natural baselines.
\end{enumerate}

\section{PROBLEM STATEMENT} 
\label{sec:problem}

We consider the problem of sequentially maximizing an unknown function $f:\mathcal{D} \rightarrow \mathbb{R}$ over a set of actions $\mathcal{D} = \mathcal{D_S} \times \mathcal{D_X}$ while satisfying a given safety constraint with respect to another unknown function $g: \mathcal{D} \rightarrow \mathbb{R}$, where $\mathcal{D_X} \subset \mathbb{R}^d$ is a compact set and $\mathcal{D_S} = [0,1]$. The function $g$ is assumed to increase monotonically in the first argument $s \in \dDS$. 

At each round $t$, the algorithm selects an action $(s_t, {\mathbf{x}}_t) \in \mathcal{D_S} \times \mathcal{D_X}$, and subsequently observes a noisy evaluation of the objective function $y^f_t = f(s_t, {\mathbf{x}}_t) + \epsilon^f_t $, as well as that of the safety function, $y^g_t = g(s_t, {\mathbf{x}}_t) + \epsilon^g_t $.  At round $t$, the selected action is a function of the history $\mathcal{H}_{t-1} = \{(s_k, {\mathbf{x}}_k, y^f_k, y^g_k) : k = 1, \dotsc, t-1\}$, and is required to satisfy the safety condition\footnote{Several existing works instead require $g(s_t, \mathbf{x}_t) \geq h$; but this is inconsequential because $g(\cdot)$ and $h$ can both simply be replaced by their negations.} $g(s_t, \mathbf{x}_t) \leq h$ as formalized below.

\paragraph{Goal:} We consider two distinct objectives: (i) finding the global safe optimum, or (ii) finding the optimal $f$-value for every $\bx \in \mathcal{D_X}$. In both cases, an algorithm must satisfy the safety constraint: (iii) $g(s_t, \mathbf{x}_t) \leq h \; \forall t \geq 1$ with high probability.

When the goal is to sequentially maximize $f$ over $\dD$ (goal (i)), we consider the following definition of cumulative regret:
\begin{equation}
\label{eqn:cum_regret_i}
    R_T = \sum_{t=1}^T r_t \text{, with } r_t = f(s_*, \bx_*) - f(s_t, \mathbf{x}_t),
\end{equation}
where $(s_*, \bx_*) = \argmax_{(s,\bx) \in \mathcal{D} : g(s,\bx) \leq h} f(s,\bx)$ is an optimal safe action.  This matches the standard cumulative regret notion in black-box optimization, but restricted to safe actions.

When the goal is to find the optimal $f$-value for every $\bx \in \dDX$, we consider the following modified definition:
\begin{equation}
\label{eqn:cum_regret_ii}
    R_T' = \sum_{t=1}^T r_t' \text{, with } r_t' = f(s^{(\bx_t)}_*, \bx_t) - f(s_t, \mathbf{x}_t),
\end{equation}
where $s^{(\bx)}_* = \argmax_{s \in \dDS : g(s,\bx) \leq h} f(s,\bx)$ denotes the optimal safe $s$ given $\bx$. Note that this definition varies from the usual notion of regret used in bandit problems. We use this formulation here to evaluate whether an algorithm achieves sublinear regret with respect to whichever $(s_t, \bx_t)$ it chooses in each round $t$. However, minimizing this quantity alone would not result in a complete evaluation of the algorithm for this goal. This is because, for example, it may choose to select the same $\bx$ in every round, and minimize regret only with respect to this $\bx$. Given that we want the algorithm to simultaneously find the maximum $f$-value for every $\bx$ for goal (ii), we additionally specify the following quantity that we also seek to minimize:
\begin{equation}
\label{eqn:dist_from_opt}
    R_T^{\mathcal{X}} = \sum_{t=1}^T r_t^{\mathcal{X}} \text{, with } r_t^{\mathcal{X}}=\max_{\bx \in \dDX} \Big(f(s^{(\bx)}_*, \bx) - f(\hat{s}^{(\bx)}_t, \bx)\Big),
\end{equation}
where $\hat{s}^{(\bx)}_t$ denotes the algorithm's ``best guess'' of the optimal safe $s \in \dDS$ for a given $\bx$ after round $t$ (see Section \ref{sec:algo} for our specific choice of $\hat{s}^{(\bx)}_t$).  

While minimizing $R_T'$ ensures that the algorithm makes progressively better choices, minimizing $R_T^{\mathcal{X}}$ ensures that for every $\bx \in \dDX$, the $f$-values of the best actions estimated by the algorithm get progressively closer to the true optima. Note that neither of these two objectives implies the other, since one specifically concerns the \emph{actions selected} ($R_T'$), while the other only evaluates the current \emph{best estimates} ($R_T^{\mathcal{X}}$). Intuitively, simultaneous minimization of both implies that not only does the algorithm get better at \textit{estimating the optimal action} for every $\bx$ (which may be achieved with pure exploration as well), it also does so by \textit{making progressively better choices} (thus trading between exploration and exploitation). 

\paragraph{Assumptions:} We adopt the standard assumption that $f$ and $g$ have bounded norm in the reproducing kernel Hilbert space (RKHS) of functions $\mathcal{D} \rightarrow \mathbb{R}$, with positive semi-definite kernel functions $k_f, k_g: \mathcal{D} \times \mathcal{D} \rightarrow \mathbb{R}$ respectively, where $\mathcal{D} = \mathcal{D_S} \times \mathcal{D_X}$. We denote the RKHS by $\mathcal{H}_{k_f}(\mathcal{D})$, and its inner product by $\langle f, k_f( (s,\bx), \cdot)\rangle_{k_f}$, and the RKHS norm by ${\vert\vert f \vert\vert}_{k_f} = \sqrt{{\langle f,f \rangle}_{k_f}}$.

To capture the smoothness of $f$, we assume a known upper bound $B_f$ on the RKHS norm of the unknown target function, i.e., $||f||_{k_f} \leq B_f$. Similarly, we assume a known upper bound $B_g$ for the safety function $g$, i.e., $||g||_{k_g} \leq B_g$. We also adopt the standard assumption of bounded variance: $k_f( (s,\bx), (s,\bx)), k_g((s,\bx), (s,\bx)) \leq 1 \; \forall (s,\bx) \in \mathcal{D}$. 

In addition, similar to \citet{losalka2023benefits}, we make the following assumptions regarding the function domain, monotonicity, and safety:
\begin{enumerate}
    \item $\mathcal{D_S} = [0,1]$ is continuous, while $\mathcal{D_X}$ can be either discrete or continuous;
    \item the function $g$ is monotonically increasing in the first argument, i.e., for all $\mathbf{x} \in \mathcal{D_X}$, $g(s,\mathbf{x})$ is an increasing function of $s \in \mathcal{D_S}$;
    \item the action $(0, \mathbf{x})$ is safe for every $\mathbf{x}$ in the domain, i.e., for all $\mathbf{x} \in \mathcal{D_X}$, $g(0,\mathbf{x}) \leq h$.
\end{enumerate}
Our algorithms can easily be adapted to avoid the third assumption, as long as we have access to an initial safe seed set. However, assuming $s=0$ to be safe allows us to measure regret with respect to the \emph{global} safe maximizer, rather than restricting to a ``reachable'' set.  We believe this assumption is natural, as $s=0$ is the most cautious choice for any $\bx$.

To derive meaningful regret bounds in our setting, it turns out to be useful to impose bounds on the \emph{maximum} growth of $f(\cdot,\bx)$ for fixed $\bx$, as well as the \emph{minimum} growth of $g(\cdot,\bx)$ for fixed $\bx$.  (In Appendix \ref{app:stuck}, we argue that such requirements cannot be avoided in general.)  Accordingly, we define $L_f > 0$ and $L'_g > 0$ to be the corresponding bounds on the growth rates, so that $\forall \bx \in \dDX, \forall s' < s$, 
\begin{align} 
    & f(s,\bx) - f(s',\bx) \leq L_f |s - s'|, \text{ and} \label{eqn:lf}  \\ 
    & g(s,\bx) - g(s',\bx) \ge L_g'|s-s'|. \label{eqn:lg'}
\end{align}
We will consider algorithms that know $L_f$ and $L'_g$; trivially, any upper bound on the former or lower bound on the latter also remains valid.  The existence of $L_f$ is a milder assumption than having a global Lipschitz constant, because it only concerns the growth with respect to $s$ (known global Lipschitz constants are common in existing algorithms such as \safeoptns, \safeoptmc and \stageoptns). 

Lastly, we make the standard assumption that the noise sequence $\{\epsilon^f_t\}_{t \geq 1}$ is conditionally $R_f$-sub-Gaussian for a fixed constant ${R_f} \geq 0$, i.e., 
\begin{equation}
    \forall t \geq 0, \forall \lambda_f \in \mathbb{R}, \mathbb{E}\left[e^{\lambda_f \epsilon^f_t} \vert \mathcal{F}_{t-1}\right] \leq \exp\left(\frac{\lambda_f^2 {R^2_f}}{2}\right),
\end{equation}
where $\mathcal{F}_{t-1}$ is the $\sigma$-algebra generated by the random variables $\{s_k, \mathbf{x}_k, \epsilon^f_k\}_{k=1}^{t-1}$ and $\mathbf{x}_t$ (similarly, $\{\epsilon^g_t\}_{t \geq 1}$ is $R_g$-sub-Gaussian). 

\section{PROPOSED ALGORITHMS} \label{sec:algo}
\paragraph{Gaussian Process Model: } Our RKHS modeling assumption naturally lends itself to the use of Gaussian process (GP) methods.  Specifically, our algorithms use the zero-mean GP models $\GP(0, k_f)$ and $\GP(0, k_g)$ for $f$ and $g$, along with an associated noise variance parameters $\lambda_f, \lambda_g > 0$ (which may differ from $R_f, R_g$).

For both $f$ and $g$, upon observing $t$ noisy values $y_1,\dotsc,y_t$, the associated posterior update equations are:
\begin{align}
    \mu_t(s, \bx) & = k_t(s, \bx)^T\left(K_t+\lambda I\right)^{-1} \mathbf{y}_{t}, \\
    k_t( (s, \bx), (s', \bx') ) & = k\left((s, \bx), (s', \bx')\right) \nonumber \\ 
    - k_t&(s, \bx)^T(K_t+\lambda I)^{-1} k_t(s', \bx'), \\
    \sigma_t^2(s, \bx) &=k_t((s, \bx), (s, \bx)), 
\end{align}
where $\mathbf{y}_t = [y_i]_{i \le t}$, $k_t(\cdot)=[k((s_i, \bx_i), \cdot)]_{i \le t}$, $K_t = [k((s_i, \bx_i),(s_j, \bx_j))]_{i,j \leq t}$, $k \in \{k_f,k_g\}$.  When considering $f$ and $g$ we suitably substitute $k \in \{k_f,k_g\}$, $\lambda \in \{\lambda_f,\lambda_g\}$, and $y_i \in \{y_i^f,y_i^g\}$ respectively.

\paragraph{Algorithm Design: } As discussed in Section \ref{sec:problem}, we consider multiple goals for our algorithms. In this section, we first outline the general structure of our proposed algorithms, and subsequently describe each specific algorithm in further detail. 

The key ideas behind our algorithms are to (i) eliminate suboptimal $\bx$'s in every round, (ii) limit the expansion of the safe set only to the regions where a ``better'' action may be found, and (iii) use an acquisition function that tries to reduce uncertainty in the set of actions that could either help expand the safe set or maximize the objective function. Ideas (i) and (ii) exploit the monotonicity of $g$ in $s$, and distinguish our algorithm from existing ones such as \safeopt (which uses idea (iii) but not (i)--(ii)).

Our algorithms use confidence bounds of the following standard form:
\begin{align}
    \mathrm{UCB}^f_{t-1}(s, \mathbf{x}) &= \mu^f_{t-1}(s, \mathbf{x})  + \beta^f_t \sigma^f_{t-1}(s, \mathbf{x}), \label{eq:ucb_f} \\ 
    \mathrm{LCB}^f_{t-1}(s, \mathbf{x}) &= \mu^f_{t-1}(s, \mathbf{x})  - \beta^f_t \sigma^f_{t-1}(s, \mathbf{x}), \label{eq:lcb_f}  
\end{align}
where $\mathrm{UCB}$ and $\mathrm{LCB}$ denote the upper and lower confidence bound respectively, $\beta^f_t$ is a time-dependent constant (and analogously with $g$ in place of $f$).  We keep $\beta^f_t$ (and $\beta^g_t$) generic here, but consider the UCB and LCB providing high-probability upper and lower bounds on $f$ (and $g$).  See Section \ref{sec:theory} for specific choices.

In each round $t=1,\dotsc,T$, the main steps of our algorithms are as follows.
\begin{enumerate}
    \item Determine the set of actions $S_t$ that can currently be classified as safe with high probability:
    \begin{equation}
    \begin{aligned}
    \label{eqn:S_t}
        S_t = \big\{(s,\bx) \in \dD &: {{\UCB}}^g_{t-1}(s,\bx) \leq h\big\} \\ &  \qquad \cup \big\{(0,\bx) : \bx \in \dDX\big\}.
    \end{aligned}
    \end{equation}
    \item Reduce the domain of $\bx$'s under consideration by eliminating any $\bx \in \mathcal{D}_\mathcal{X}^{t-1}$ that satisfies an elimination criteria, $\mathtt{elim}_t(\bx) = \texttt{true}$ (detailed below) to form the set $\mathcal{D}_\mathcal{X}^t$ (starting with $\mathcal{D}_\mathcal{X}^0 = \dDX$):
    \begin{equation}
    \label{eqn:DXt_elim}
        \mathcal{D}_\mathcal{X}^{t} = \mathcal{D}_\mathcal{X}^{t-1} \setminus \left\{\bx \in \mathcal{D}_\mathcal{X}^{t-1}: \mathtt{elim}_t(\bx) = \texttt{true} \right\}.
    \end{equation}
    \item For every $\bx \in \mathcal{D}_\mathcal{X}^t$, find the $s$-values on the current ``safe boundary'', given by the highest $s \in \dDS$ such that safety is guaranteed with high probability (denoted by $s^{(\bx)}_t$). Form the set $G_t$ with these $(s^{(\bx)}_t, \bx)$ pairs if there is a possibility of expansion to a more optimal $f$-value (as decided by a function $\mathtt{expd}_t(s,\bx)$).  Mathematically, we have:
    \begin{align}
        s^{(\bx)}_t  &= \max \big\{s \in \dDS : (s,\bx) \in S_t\big\},    \label{eqn:s_t^x} \\ 
        G_t = \big\{&(s^{(\bx)}_t, \bx) :\, \bx \in \mathcal{D}_\mathcal{X}^t ,\, \mathtt{expd}_t(s,\bx) = \texttt{true} \label{eqn:G_t}\big\}.
    \end{align}
    \item For every $\bx \in \mathcal{D}_\mathcal{X}^t$, find a safe action that maximizes $\mathrm{UCB}^f_{t-1}(s,\bx)$, and form the set of such maximizers, $M_t$, as follows:
    \begin{align}
        \hat{s}^{(\bx)}_t &= \argmax_{s\in\dDS:s \leq s^{(\bx)}_t} \mathrm{UCB}_{t-1}^f(s,\bx), \label{eqn:hat_s_t^x}\\ 
        M_t &= \big\{(\hat{s}^{(\bx)}_t, \bx) : \bx \in \mathcal{D}_\mathcal{X}^t\big\}. \label{eqn:M_t}
    \end{align}
    Note that we use the same notation as in \eqref{eqn:dist_from_opt} here (i.e.,  $\hat{s}^{(\bx)}_t$), because our proposed algorithms use the maximizer (over $s$) of $\mathrm{UCB}^f_{t-1}(s,\bx)$ as the current estimate of the optimal safe $s$ for a given $\bx$.
    \item Use an acquisition function $\mathtt{acq}_t(s,\bx)$ to select an action as follows: 
    \begin{equation}
    \label{eqn:acq}
        (s_t, \bx_t) = \argmax_{(s,\bx)} \, \mathtt{acq}_t(s, \bx).
    \end{equation}
\end{enumerate}

Next, we describe the three key functions, $\mathtt{elim}_t, \mathtt{expd}_t$ and $\mathtt{acq}_t$, which vary corresponding to the different problem settings as discussed earlier. 

\begin{algorithm}[!t]
\caption{\msafeopt}
\label{alg:struct}
\begin{algorithmic}[1]
\vspace{.25em}
\State \textbf{Input:} Prior $\GP(0, k_f)$, $\GP(0, k_g)$, parameters $\lambda_f, \lambda_g, L_f, L_g',  \{\beta^f_t\}_{t \ge 1},  \{\beta^g_t\}_{t \ge 1}$
\State $\mathcal{D}^0_\mathcal{X} = \mathcal{D_X}$
\For{$t = 1, \dotsc, T$}
    \State $\mathcal{D}^{t}_\mathcal{X} = \mathcal{D}^{t-1}_\mathcal{X} \setminus \{ 
\bx \in \mathcal{D}^{t-1}_\mathcal{X} \,:\, \mathtt{elim}_t(\bx) = \texttt{true}  \}$ 
    \State \Comment{{\small eliminate all suboptimal $\bx$ to form $\mathcal{D}^{t}_\mathcal{X}$}} 
    \State $G_t = \emptyset$
    \State $M_t = \emptyset$
    \For{$\bx \in \mathcal{D}^t_\mathcal{X}$}   \Comment{{\small find max. safe $s \, \forall \bx \in \mathcal{D}^t_\mathcal{X}$}}
            \If{$ {{\UCB}}^g_{t-1}(s, \bx) > h \,\forall s \in \dDS$}
                \State $s_t^{(\bx)} = 0$
            \ElsIf{$ \exists s\in \dDS : {{\UCB}}^g_{t-1}(s, \bx) = h $}
                \State{$
                \begin{aligned}
                    s_t^{(\bx)}& = \max\{s \in \dDS, {{\UCB}}^g_{t-1}(s, \bx) = h \} \\
                \end{aligned}
                $}
            \Else
                \State $s_t^{(\bx)} = 1$
            \EndIf            
        \If{$\mathtt{expd}_t(s_t^{(\bx)},\bx)$ is \texttt{true}}  \Comment{{\small form $G_t$}} 
            \State $G_t = G_t \cup \{ (s_t^{(\bx)}, \bx)\}$
        \EndIf
        
    \State $\hat{s}_t^{(\bx)} = \argmax_{s \in \dDS:s \leq s_t^{(\bx)}} \mathrm{UCB}^f_{t-1}(s,\bx)$
    \State $M_t = M_t \cup \{ (\hat{s}_t^{(\bx)}, \bx)\}$  \Comment{{\small form $M_t$}}  
\EndFor
\State $(s_t, \bx_t) \; = \; \argmax_{(s,\bx):\bx \in \mathcal{D}^{t}_\mathcal{X}} \,\, \mathtt{acq}_t(s, \bx)$ 
\State Update posterior to get $\mu^f_t, \sigma^f_t, \mu^g_t, \sigma^g_t$
\EndFor
\end{algorithmic}
\end{algorithm}

\paragraph{Case 1: Maximizing $f$ across $\dD$:} 

We first consider the most standard cumulative regret notion, corresponding to $R_T$ in \eqref{eqn:cum_regret_i}.  
In this case, the three functions in Algorithm \ref{alg:struct} are defined as follows:
\begin{itemize}
    \item $\mathtt{elim}_t: $ We eliminate $\bx$'s that are suboptimal according to the confidence bounds. For defining suboptimality, we first define the following term: 
        \begin{equation}
        \label{eqn:us_t^x}
            \begin{aligned}
            \usxt & = \max \big\{s\in \dDS : \\
            &\quad \mathrm{LCB}_{t-1}^g(s^{(\bx)}_t,\bx) + L_g'|s-s^{(\bx)}_t| \leq h\big\}.
            \end{aligned}
        \end{equation}
    $\usxt$ essentially indicates the highest $s \in \dDS$ for which the safety function could be at most $h$ optimistically (since even with the minimum growth rate, $g$ exceeds $h$ beyond $\usxt$). 
    
    Suboptimality of $\bx$ is decided based on the following two conditions: (i) the highest $\UCB^f_{t-1}$ for the $(s,\bx)$'s currently known to be safe is lower than the $\LCB^f_{t-1}$-value of some $(s',\bx')$ known to be safe, \textit{and} (ii) the maximum possible $f$-value at $(\usxt, \bx)$ is less than the $\LCB^f_{t-1}$-value of some $(s',\bx')$ known to be safe. Thus, $\mathtt{elim}_t(\bx)$ is \texttt{true} if both of the following conditions hold:
    \begin{align}
        & \text{(i) } \max_{s \leq s^{(\bx)}_t} \{\UCB^f_{t-1}(s,\bx)\} \nonumber \\ 
        &\qquad \qquad < \max_{(s',\bx') \in S_t}\big\{\LCB^f_{t-1}(s',\bx')\big\},    \label{eqn:elim_i_i}\\ 
        & \text{(ii) } \UCB^f_{t-1}( s^{(\bx)}_t,\bx) + L_f|\usxt - s^{(\bx)}_t| \nonumber  \\ 
        & \qquad \qquad \leq \max_{(s',\bx') \in S_t}\big\{\LCB^f_{t-1}(s',\bx')\big\}.\label{eqn:elim_i_ii}
    \end{align}
    These criteria ensure that even when being optimistic, any safe action corresponding to $\bx$ (either within the current safe region as in \eqref{eqn:elim_i_i}, or after expanding further as in \eqref{eqn:elim_i_ii}) cannot result in a higher $f$-value than $f(s',\bx')$.
    \item ${\mathtt{expd}_t: }$ We include an action $({s}^{(\bx)}_t, \bx)$ in the set $G_t$ only if expanding to $\underline{s}^{(\bx)}_t$ could optimistically lead to a better $f$-value than the one currently found. Thus, $\mathtt{expd}_t(\bx)$ is set to \texttt{true} if the following condition holds:
    \begin{align}
    \label{eqn:exp_i}
        \UCB^f_{t-1}({s}^{(\bx)}_t,&\, \bx) + L_f|\underline{s}^{(\bx)}_t-{s}^{(\bx)}_t| \nonumber \\ 
        & > \max_{(s',\bx') \in S_t}\big\{\LCB^f_{t-1}(s',\bx')\big\}.
    \end{align}
    \item ${\mathtt{acq}_t: }$ We define the acquisition function as:
    \begin{align}
    \label{eqn:acq_i}
        & \mathtt{acq}_t(s,\bx) \\
        & \, = \begin{cases}
             \max \big\{\beta^f_t\sigma^f_{t-1}(s,\bx), \beta^g_t\sigma^g_{t-1}(s,\bx) \big\},  \\ 
             \qquad \qquad \qquad \qquad \qquad \qquad \quad \, \text{if } (s,\bx) \in G_t;  \\ 
            \beta^f_t\sigma^f_{t-1}(s,\bx), \text{ if } (s,\bx) \in M_t \text{ and } (s,\bx) \notin G_t; \\ 
            0, \text{ otherwise,}
        \end{cases}          \nonumber
    \end{align}
    in order to reduce uncertainty of $f$ within the potential maximizers $M_t$, and reduce the uncertainty of \emph{both $f$ and $g$} in the expander set $G_t$.
\end{itemize}

\paragraph{Case 2: Maximizing $f$ for every $\bx \in \dDX$:} Suppose that the goal is to find the optimal safe $s$ for \emph{every} $\bx$ (goal (ii) in Section \ref{sec:problem}). In this case, we modify $\mathtt{elim}_t$ and $\mathtt{expd}_t$ as follows, while keeping $\mathtt{acq}_t$ unchanged.

\begin{itemize}
    \item $\mathtt{elim}_t: $ Since we want to find the optimal action corresponding to \textit{every} $\bx$, we should not eliminate any $\bx$'s from consideration in this case. Thus, we define $\mathtt{elim}_t(\bx) = \texttt{false} \,\, \forall \bx \in \dDX$.
    
    \item ${\mathtt{expd}_t: }$ With a similar motivation to Case 1, $\mathtt{expd}_t(\bx)$ is set to \texttt{true} if the following holds:
    \begin{align}
    \label{eqn:exp_ii}
        \UCB^f_{t-1}(& {s}^{(\bx)}_t, \bx) + L_f|\underline{s}^{(\bx)}_t -{s}^{(\bx)}_t| \nonumber \\ 
        & \qquad > \max_{s\leq {s}^{(\bx)}_t}\{\LCB^f_{t-1}(s,\bx)\},
    \end{align}
    where $\underline{s}^{(\bx)}_t$ is the highest ``potentially safe'' $s$ as earlier (see \eqref{eqn:us_t^x}).  The condition \eqref{eqn:exp_ii} states that the optimistic $f$-value at $\underline{s}^{(\bx)}_t$ is better than a pessimistic value among the $s$ known to be safe given $\bx$.  Note that unlike Case 1, we use $\bx$ on both sides and do \emph{not} compare to any $\bx' \ne \bx$; this is because here we must find the best $s$ for \emph{every} $\bx$.
    
\end{itemize}

\paragraph{Case 3: Both $f$ and $g$ are monotone: } Here we consider the scenario where both $f$ and $g$ are monotonically increasing functions in $s$.  As a first sub-case, we again consider the goal \eqref{eqn:cum_regret_i} associated with finding a global safe maximizer.  
While the same algorithm as case 1 would work here as well, we find that the algorithm can be simplified substantially. Since we know that the optimal $f$-value will be found along the safe boundary, there is no requirement to separately maintain the set $M_t$. 

We simplify the three key functions in Algorithm \ref{alg:struct} as follows. First, for defining $\mathtt{elim}_t$, it suffices to consider \eqref{eqn:elim_i_ii} only. Next, we always expand for any $\bx$ that is not eliminated, i.e., $\forall \bx \in \mathcal{D}^t_{\mathcal{X}}$, $\mathtt{expd}_t$ is set to \texttt{true}. Finally, the acquisition function is simplified as follows:
\begin{align}
\label{eqn:acq_iii}
        & \mathtt{acq}_t(s,\bx) \\
        & \, = \begin{cases}
             \max \big\{\beta^f_t\sigma^f_{t-1}(s,\bx), \beta^g_t\sigma^g_{t-1}(s,\bx) \big\}, \text{ if } (s,\bx) \in G_t; \\ 
             0, \text{ otherwise.} \nonumber
        \end{cases}
\end{align}

As a second sub-case, with \emph{both} $f$ and $g$ still being monotone, we may consider the task of finding the best (highest) $s$ for every $\bx$.  In this scenario, removing the elimination step altogether, setting $\mathtt{expd}_t(\bx) = \mathtt{false}$ when $s^{(\bx)}_t = 1,$ and modifying the acquisition function to only consider $\sigma_{t-1}^g$ essentially results in the \msafeucb algorithm of \citet{losalka2023benefits}. In Appendix \ref{sec:app_msafeucb}, we show how their theoretical guarantees can be recovered as a special case of ours. 

\paragraph{Extensions: } In Appendix \ref{app:extensions}, we outline several straightforward extensions of our algorithm, including multiple safety functions, joint RKHS modeling of $(f,g)$, and the presence of contextual variables.

\section{THEORETICAL RESULTS} \label{sec:theory}
Our theoretical analysis relies on the widely-used notion of \emph{information gain}, which we define separately for $f$ and $g$ as follows: 
\begin{align}
    \gamma^f_t := \max_{A \subset \mathcal{D}: |A| = t} I(y^f_A; f_A), \\
    \gamma^g_t := \max_{A \subset \mathcal{D}: |A| = t} I(y^g_A; g_A),
\end{align}
where $I(y^f_A; f_A)$ denotes the mutual information between $f_A = [f(s, \bx)]_{(s, \bx) \in A}$ and $y^f_A = f_A + \epsilon^f_A$, and where $\epsilon^f_A \sim \mathcal{N}(0, \lambda_f I)$ (similarly, for $I(y^g_A; g_A)$).  The information gain essentially captures the amount of uncertainty reduction in the function as a result of observing $t$ noisy evaluations.

Recall that our confidence bounds \eqref{eq:ucb_f}--\eqref{eq:lcb_f} depend on parameters $\beta_t^f$ and $\beta_t^g$ that have been generic until now.  We will state our results keeping them generic, but requiring their validity; formally, we say that $(\beta_t^f,\beta_t^g)$ provide \emph{$(1-\delta)$-valid confidence bounds} if, with probability at least $1-\delta$, for all $(s, \bx) \in \mathcal{D}, t \geq 1$,
\begin{align}
    | \mu^f_{t-1} (s, \bx) - f(s, \bx)| & \leq  \beta_t \sigma^f_{t-1}(s, \bx) \label{eq:valid_conf_f} \text{, and} \\ 
    | \mu^g_{t-1} (s, \bx) - g(s, \bx)| & \leq  \beta_t \sigma^g_{t-1}(s, \bx). \label{eq:valid_conf_g}
\end{align}
Additionally, our proofs rely on the $\beta_t$-terms being non-decreasing. The following lemma from \citet{chowdhury2017kernelized} (namely, their Theorem 2 and a union bound over $f$ and $g$) provides a well-known such choice of $(\beta_t^f,\beta_t^g)$; we will also mention an alternative below.

\begin{lemma}
\label{lemma:beta}
For any $\delta > 0$, the parameters
\begin{align}
&\beta^f_t = B_f+{R_f}\sqrt{2(\gamma^f_{t-1}+1+\ln{(2/ \delta)})}, \text{ and} \\
&\beta^g_t = B_g+{R_g}\sqrt{2(\gamma^g_{t-1}+1+\ln{(2/ \delta)})} \label{eq:beta_t} 
\end{align}
provide $(1-\delta)$-valid confidence bounds.
\end{lemma}

We are now ready to state our main theorems, all of which are proved in Appendix \ref{sec:proofs}. Recall that our three regret notions $R_T$, $R'_T$, and $R^{\mathcal{X}}_T$ are defined in \eqref{eqn:cum_regret_i}--\eqref{eqn:dist_from_opt}, and $L_f,L'_g$ are defined in \eqref{eqn:lf}--\eqref{eqn:lg'}. 

\begin{theorem}
\label{theorem:regret_bound}
\textit{Under the setup and assumptions of Section \ref{sec:problem} and any non-decreasing $\beta^f_t, \beta^g_t$ providing $(1-\delta)$-valid confidence bounds, Algorithm \ref{alg:struct} (in both case 1 and case 3) satisfies the following with probability at least $1-\delta$:}
\begin{align}
\label{eqn:theorem_1}
R_T = O\left( \left(1+\frac{L_f}{L'_g}\right) \left(\beta^g_T\sqrt{T\gamma^g_T} + \beta^f_T\sqrt{T\gamma^f_T} \right) \right).
\end{align}
\end{theorem}

\begin{theorem}
\label{theorem:regret_bound_R_T_ii}
\textit{Under the setup and assumptions of Section \ref{sec:problem} and any non-decreasing $\beta^f_t, \beta^g_t$ providing $(1-\delta)$-valid confidence bounds, Algorithm \ref{alg:struct} (in case 2) satisfies the following with probability at least $1-\delta$:} 
\begin{align}
\label{eqn:theorem_2}
R_T' = O\left( \left(1+\frac{L_f}{L'_g}\right) \left(\beta^g_T\sqrt{T\gamma^g_T} + \beta^f_T\sqrt{T\gamma^f_T} \right) \right).
\end{align}
\end{theorem}

\begin{theorem}
\label{theorem:regret_bound_D_T_ii}
\textit{Under the setup and assumptions of Section \ref{sec:problem}, and any non-decreasing $\beta^f_t, \beta^g_t$ providing $(1-\delta)$-valid confidence bounds, Algorithm \ref{alg:struct} (in case 2) satisfies the following with probability at least $1-\delta$:}
\begin{align}
R_T^{\mathcal{X}} = O\left( \left(1+\frac{L_f}{L'_g}\right) \left(\beta^g_T\sqrt{T\gamma^g_T} + \beta^f_T\sqrt{T\gamma^f_T} \right) \right).
\end{align}
\end{theorem}

We note that the above regret bounds can be refined to the following form:
\begin{equation}
    \label{eqn:modified_regret}
    R_T = O\left(\beta_T^f \sqrt{T \gamma_T^f} + \frac{L_f}{L_g'}\beta_T^g \sqrt{T \gamma_T^g} \right)
\end{equation}
by modifying the acquisition function as follows: 
\begin{equation}
    \label{eqn:modified_acq}
    \mathtt{acq}_t(s,\bx) = \max\{\beta^f_t \sigma^f_{t-1}(s,\bx), (L_f/L_g') \beta^g_t \sigma^g_{t-1}(s,\bx)\}
\end{equation}
if $(s,\bx) \in G_t$ (with the other cases remaining unchanged in \eqref{eqn:acq_i} and \eqref{eqn:acq_iii}). This refined bound applies to $R_T$ (in both case 1 and case 3), $R_T'$ (in case 2), and $R_T^{\mathcal{X}}$ (in case 2), and provides desirable properties with respect to the scaling of $f$ and $g$. (See Appendix \ref{sec:refined_regret} for a proof of the validity of the refined regret bounds, and a discussion on the scaling properties.) However, this comes with the trade-off of having to incorporate the constants $L_f$ and $L_g'$ into the acquisition function. This introduces additional dependence of the algorithm on these constants, which may be undesirable. Therefore, we we primarily focus on the algorithm with the earlier acquisition function \eqref{eqn:acq}, which has similar sublinear regret guarantees.

For all of the above theorems, our regret bounds have the same $\beta_T \sqrt{T \gamma_T}$ form (with $f$ or $g$ superscripts) as GP-UCB \citep{srinivas2012information} and other related algorithms (e.g., \citep{chowdhury2017kernelized}), and hence also the same bounds when applied to specific kernels.  For instance, for the squared exponential kernel, we have $\gamma_T = O({\ln^{d+1}{T}})$ \citep{srinivas2012information}, which implies sublinear regret via Lemma \ref{lemma:beta}.  

For the Mat\'ern-$\nu$ kernel, we have $\gamma_T = O\big( T^{\frac{d}{2\nu + d}} \log T \big)$ \citep{vakili2021information}.  Since $\beta_T$ is also linear in $\gamma_T$ in Lemma \ref{lemma:beta}, this only guarantees sublinear regret if $\frac{d}{2\nu + d} < \frac{1}{2}$, i.e., $\nu > \frac{d}{2}$.  Fortunately, alternative confidence bounds have recently been given that guarantee sublinear regret without this restriction \citep{whitehouse2023sublinear}, and we can directly make use of these since our theorems are stated in terms of generic confidence bounds.  The changes required for these variations are identical to the vanilla setting without safety constraints, so we do not repeat them.

\begin{samepage}
\begin{figure*}
  \centering
  \setlength\tabcolsep{2pt}
  \begin{tabular}{cc}
     \includegraphics[width=0.5\linewidth]{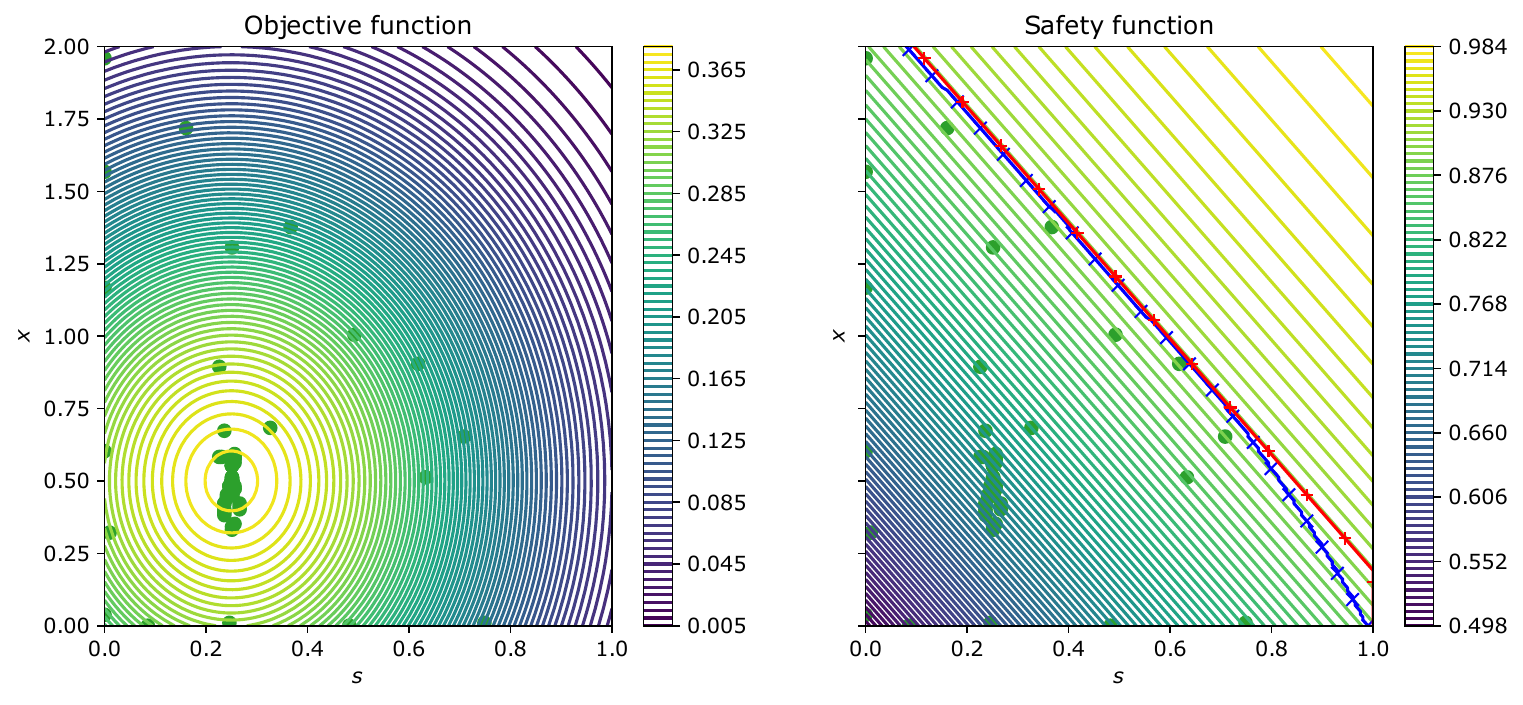} &
     \includegraphics[width=0.5\linewidth]{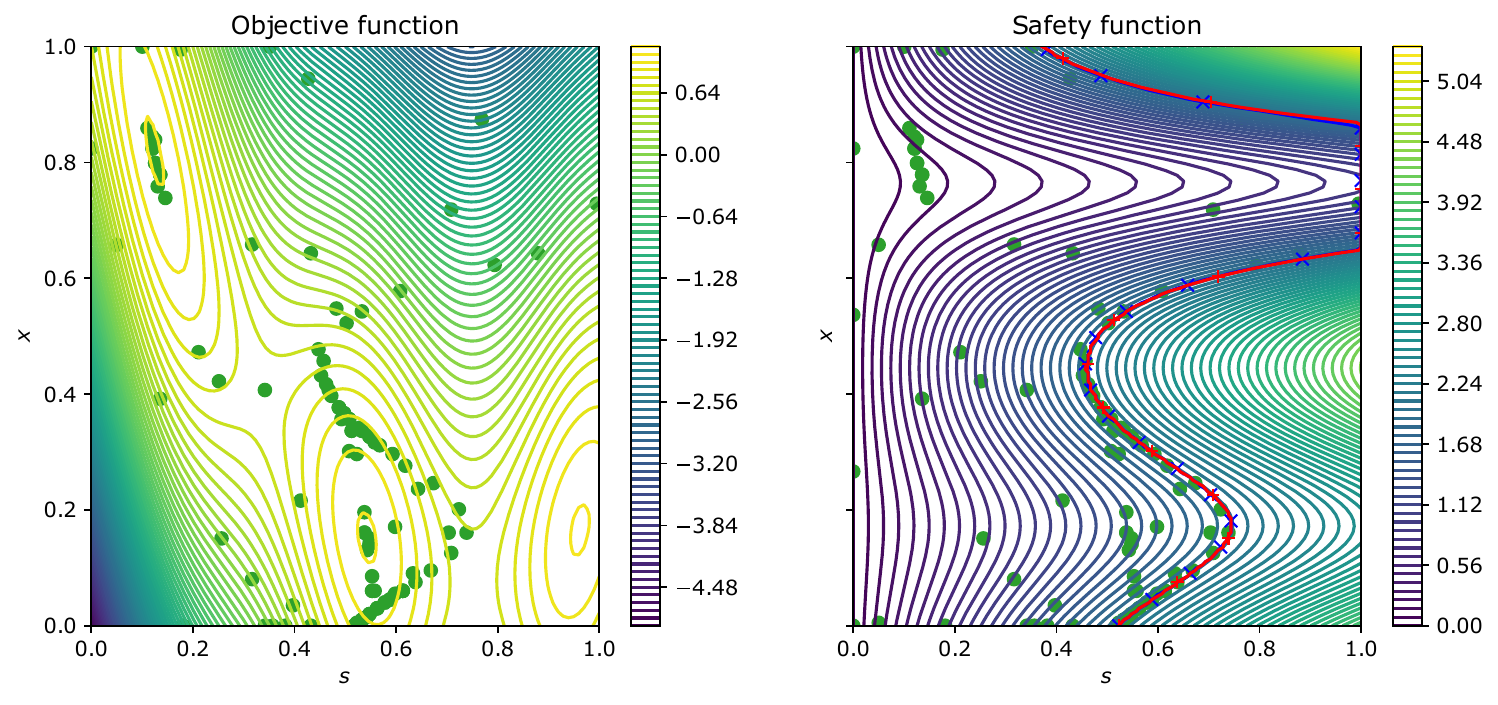} \\ 
     \includegraphics[width=0.5\linewidth]{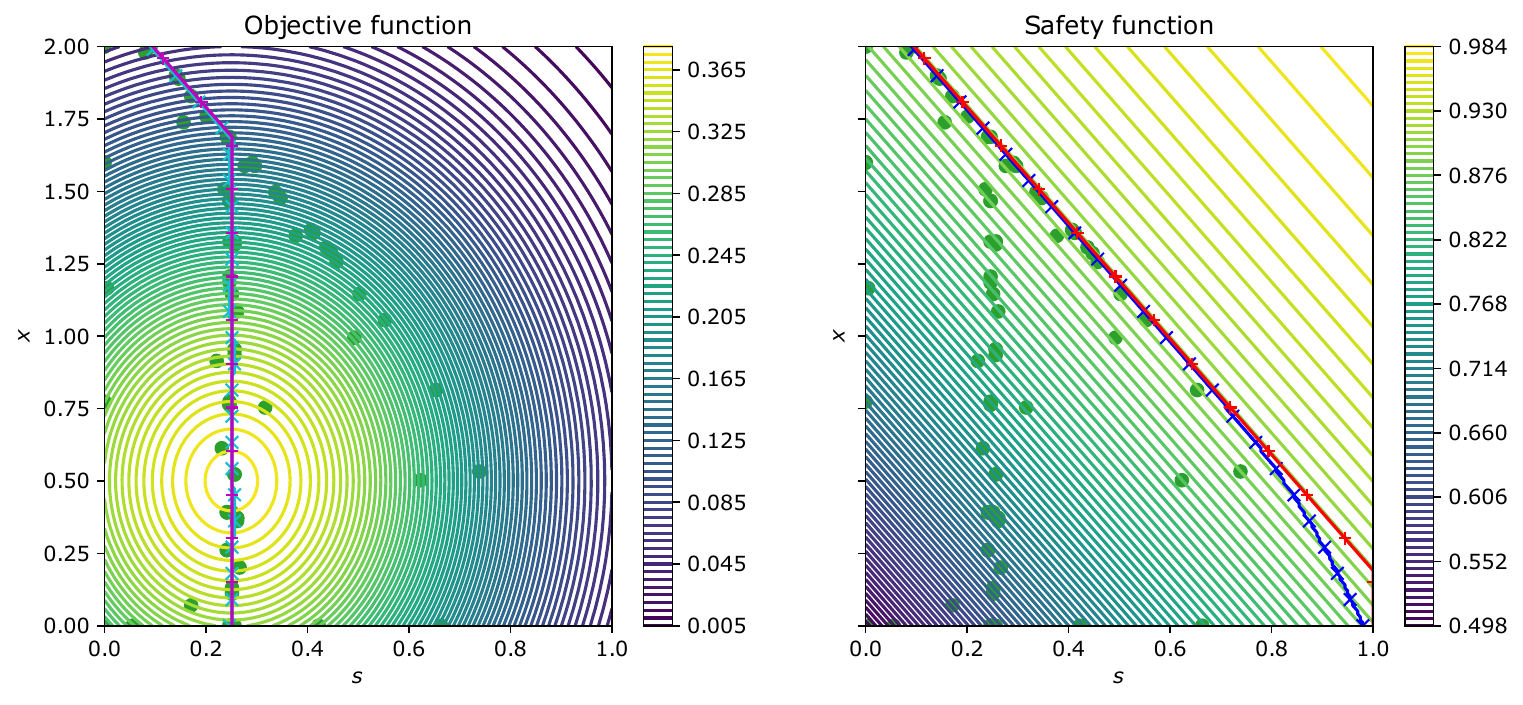} &
     \includegraphics[width=0.5\linewidth]{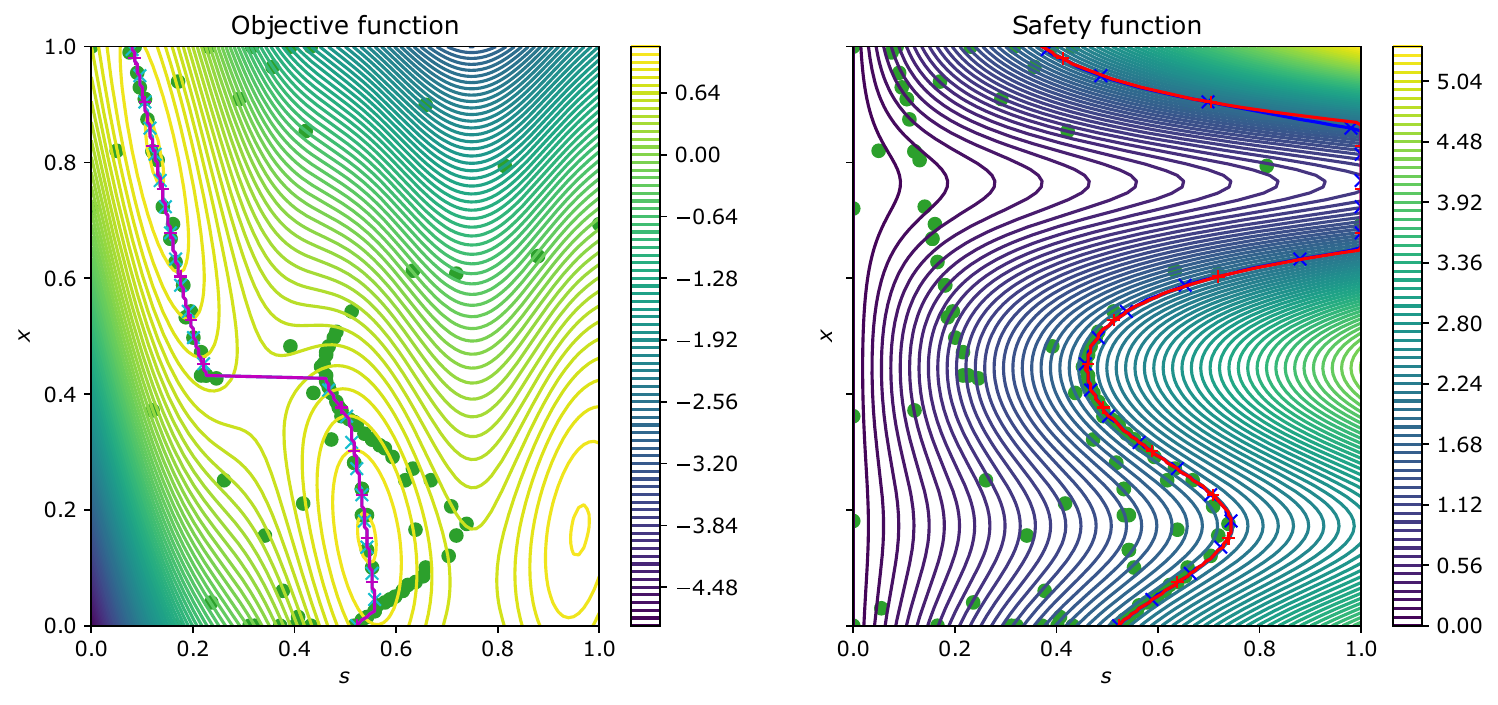} \\
     {\scriptsize Simulated clinical trial $(f_{\mathrm{eff}}, g_{\mathrm{tox}})$} & {\scriptsize Synthetic 2D functions $(f_{\mathrm{syn_1}}, g_{\mathrm{syn_1}})$}
    \end{tabular}
  \caption{Actions sampled by \msafeoptns \footnotemark[3] ~in case 1 (top row) and case 2 (bottom row), along with the safe boundaries discovered in blue and true safe boundaries in red (2\textsuperscript{nd} and 4\textsuperscript{th} column). In case 2, the 1\textsuperscript{st} and 3\textsuperscript{rd} columns also show the optimal $s$ discovered for every $\bx$ in cyan, and the true optimal $s$-values in magenta.
  \label{fig:bo_plots}}
  \bigskip\bigskip
\end{figure*}

\begin{figure*}[t!]
  \centering
  \setlength\tabcolsep{2pt}
  \begin{tabular}{ccc}
     \includegraphics[width=0.33\linewidth]{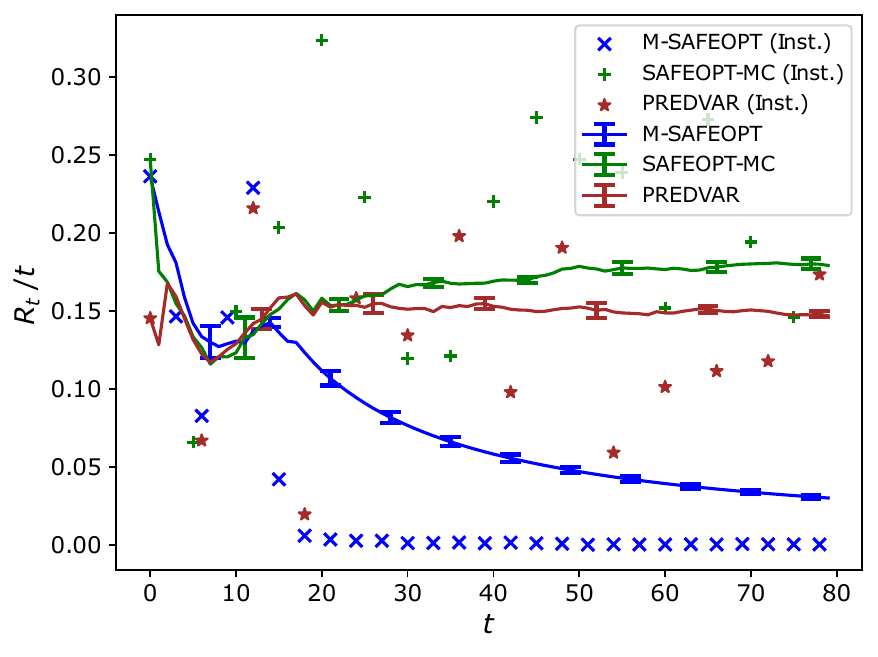} &
     \includegraphics[width=0.33\linewidth]{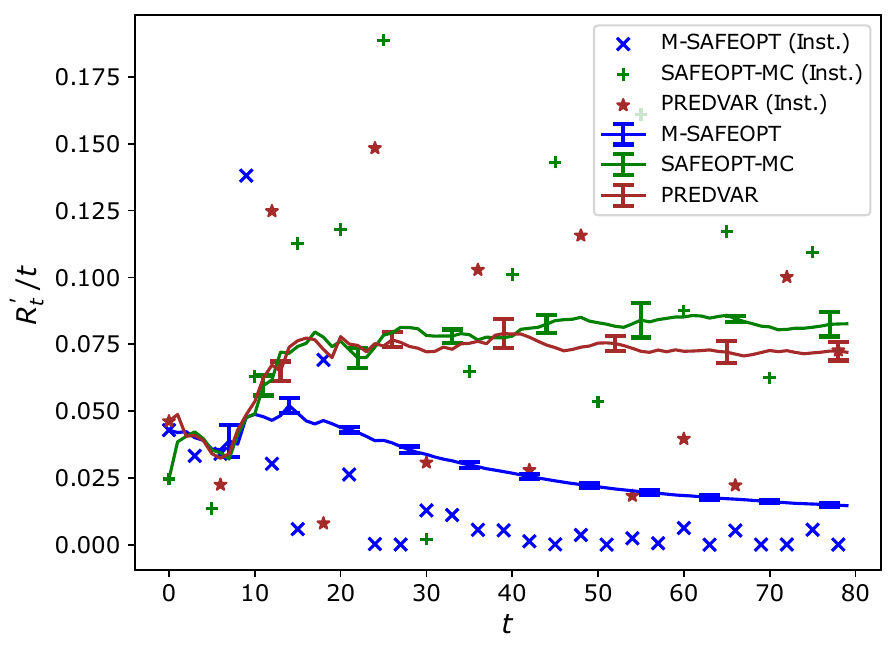} &
     \includegraphics[width=0.33\linewidth]{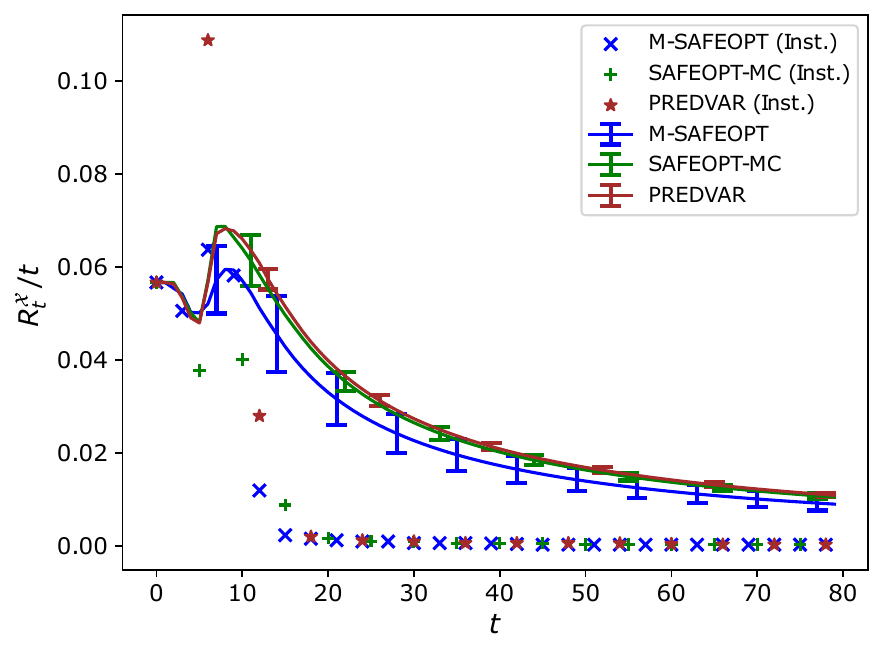} \\ 
     \includegraphics[width=0.33\linewidth]{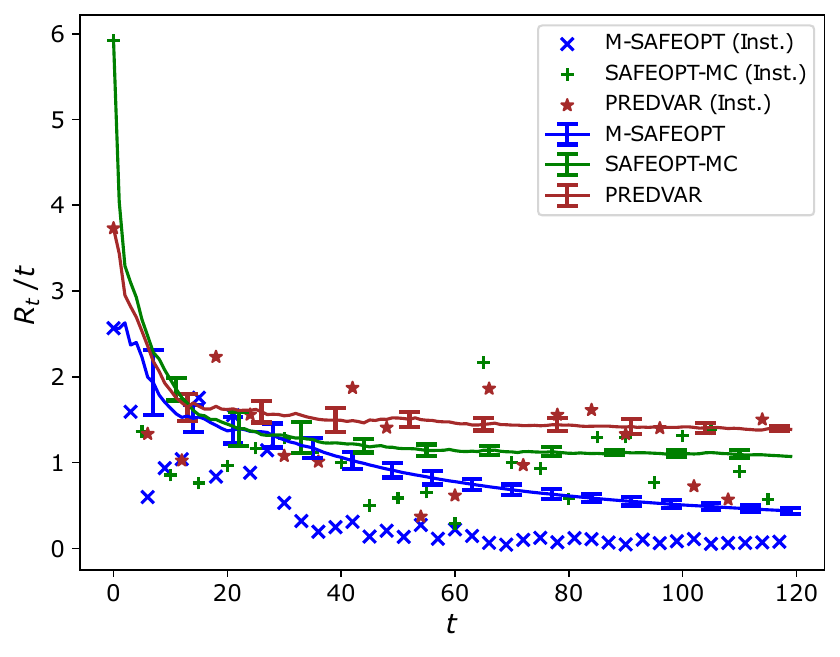} &
     \includegraphics[width=0.33\linewidth]{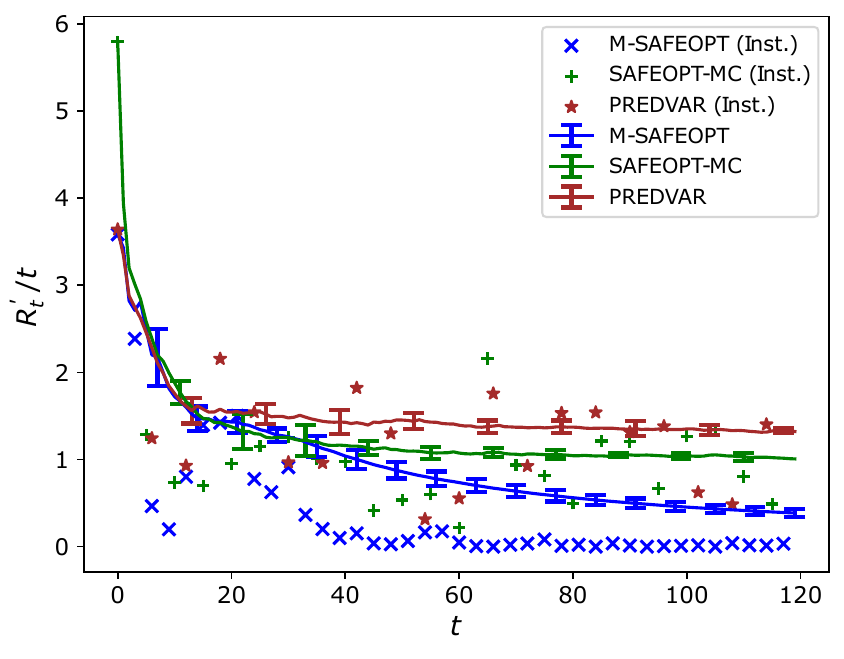} &
     \includegraphics[width=0.33\linewidth]{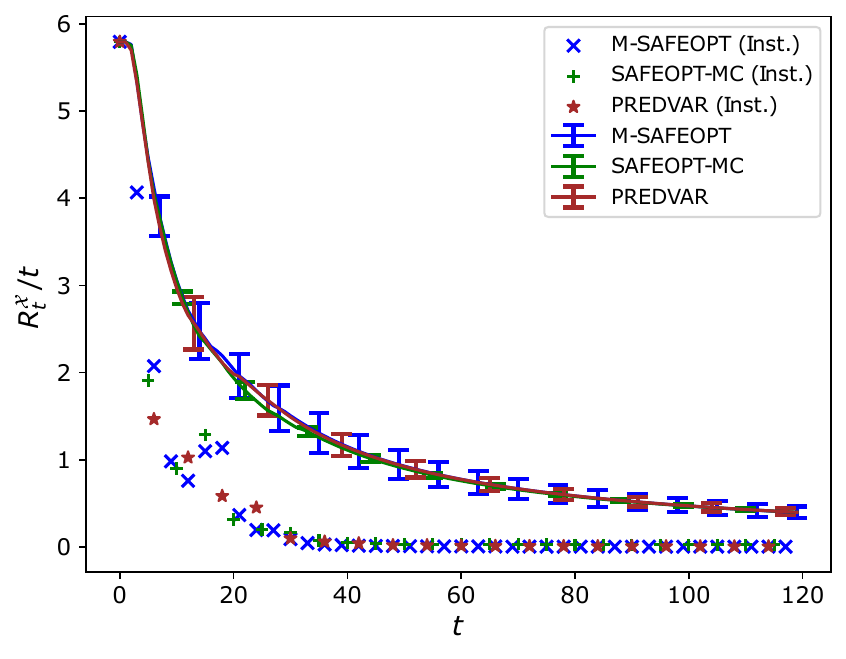}
    \end{tabular}
  \caption{The top row shows the regret plots for the simulated clinical trial experiment, and the bottom row shows that for the synthetic 2D experiment for \msafeoptns, along with baseline algorithms. The first column shows the plot for $R_t/t$ (for Case 1), while columns 2 and 3 show $R'_t/t$ and $R^{\mathcal{X}}_t/t$ (for Case 2). The corresponding instantaneous regret values are shown using markers.} 
  \label{fig:regret_plots}
\end{figure*}
\end{samepage}

\section{EXPERIMENTS} \label{sec:exp}
In this section, we present empirical results to complement our theoretical findings, by running the \msafeopt algorithm (in Case 1 and Case 2) and comparing with baseline algorithms \footnote{The code is available at \url{https://github.com/arpanlosalka/m-safeopt}.}. The primary goal of the experiments is to (i) verify that our cumulative regret notions demonstrate sublinear behavior, (ii) verify that unsafe actions are not sampled, and (iii) demonstrate performance gains over existing algorithms. We primarily use \safeoptmc \citep{berkenkamp2021bayesian} and the purely exploratory \predvar \citep{schreiter2015safe} algorithms for comparisons. \safeopt \citep{sui2015safe} and \msafeucb \citep{losalka2023benefits} are skipped, since they are designed to work with a single function $f$. 
Additional experimental results are provided in Appendix \ref{sec:app_more_results}, and the details (e.g., descriptions of baselines, choices of kernels and $\beta_t$) are given in Appendix \ref{sec:app_exp_details}.

\paragraph{Simulated Clinical Trial: } For this experiment, we use the logistic function as a model of the dose-toxicity and dose-efficacy behaviors. Specifically, following \citet{cai2014bayesian}, we use the functions 
\begin{align}
    &f_{\mathrm{eff}}(d_1, d_2) = \left\{1+e^{\theta^f_0 - \theta^f_1 d_1 - \theta^f_2 d_2 - \theta^f_3 d_1^2 - \theta^f_4 d_2^2}\right\}^{-1} \label{eqn:dose_eff} \\
    & \text{and }\quad g_{\mathrm{tox}}(d_1, d_2) = \left\{1+e^{-\theta^g_1 d_1 - \theta^g_2 d_2}\right\}^{-1}, \label{eqn:dose_tox}
\end{align}
\footnotetext[3]{See Appendix \ref{sec:app_more_results} for similar plots with the baselines.}
where $d_1 \, (s)$ and $d_2 \, (\bx)$ denote the dosage of two drugs, and $\theta_i$'s denote suitable parameters (see Appendix \ref{sec:app_exp_details} for details). While $g_{\mathrm{tox}}$ increases monotonically with the dosage of both drugs, the efficacy peaks at an intermediate dose level of the drugs and then decreases. 

\paragraph{Synthetic 2D function:} In this section, we use the scaled Branin function from \citep{picheny2013benchmark} as the objective function $f_{\mathrm{syn_1}}$. It has a more complex optimization surface, with three local optima over the domain considered, $\dD = [0,1]^2$. For the safety function $g_{\mathrm{syn_1}}$, we use a slightly modified form of $f_{\mathrm{syn_2}}$ from \citep{losalka2023benefits}, such that optimization becomes more challenging for both goal (i) (global optimization) and goal (ii) (optimization  $\forall\bx$).

\paragraph{Observations: } For both the above experiments, we observe in Figure \ref{fig:bo_plots} that unsafe actions are not sampled, and for both goal (i) and goal (ii), the samples of \msafeopt tend to be near the optimal actions instead of unnecessarily exploring suboptimal regions of the input space. 

The regret plots in Figure \ref{fig:regret_plots} demonstrate that \msafeopt achieves sublinear regret (for suitable notions of regret, depending on the goal), whereas the baseline algorithms fail to do so (except for $R^{\mathcal{X}}_T$). Both of the baseline algorithms continue to explore suboptimal regions, either near the safe boundary (\safeoptmcns) or throughout the safe region (\predvarns), and this gets reflected in the $R_t/t$ and $R'_t/t$ plots. However, they perform well with respect to $R^{\mathcal{X}}_t$, which considers only the \textit{best action discovered}, and not the \textit{action chosen} in each round for evaluation. As discussed in Section \ref{sec:problem}, our goal (ii) naturally leads to a requirement of both $R'_T$ and $R^{\mathcal{X}}_T$ being small, rather than either of them alone.

Next, we highlight some observations regarding the performance of \msafeopt based on our experiments. (See Appendix \ref{app:results_discussion} for further discussion on \safeoptmc and \predvarns.)
\begin{itemize}\itemsep0ex
    \item \msafeopt (Case 1), when used for goal (i) (as in Section \ref{sec:problem}) is able to eliminate suboptimal $\bx$'s (via $\mathtt{elim}_t$) and converge quickly to the regions in which there is a higher probability of finding the safe optimal action. For instance, this can be observed for $\bx \in [1,2]$ in row 1-column 1 of Figure \ref{fig:bo_plots} for $(f_{\mathrm{eff}}, g_{\mathrm{tox}})$ and for $\bx \in [0.9, 1]$ in row 1-column 3 for $(f_{\mathrm{syn}_1}, g_{\mathrm{syn}_1})$, where the algorithm stops exploring once it has located better actions near/at the true optimum.
    \item \msafeopt (in Case 1 and Case 2) also limits unnecessary expansion beyond the currently discovered safe boundary (via $\mathtt{expd}_t$), helping the algorithm to converge to the optimal region quicker that \safeopt. For instance, this is observed for $\bx \in [0, 0.25]$ in row 1-column 2 (Case 1) of Figure \ref{fig:bo_plots} for $(f_{\mathrm{eff}}, g_{\mathrm{tox}})$, and similarly in row 2-column 2 of the same figure (Case 2).
    \item The safe boundary eventually discovered by \msafeopt (in both Case 1 and Case 2) tends to diverge from the true safe boundary with respect to $g$. This is also due to elimination and  limiting expansion. Note that when no $\bx$ can be eliminated (in Case 2), the same phenomenon (e.g., $\bx \in [0, 0.25]$ in row 2-column 2 of Figure \ref{fig:bo_plots}) can be observed due to non-expansion of suboptimal $\bx$'s.  
    \item In Case 2 (for goal (ii)), \msafeopt is able to find the optimal $s$ for every $\bx \in \dDX$ almost exactly (as seen in Figure \ref{fig:bo_plots}, and also reflected in the regret plot in the third column of Figure \ref{fig:regret_plots} with respect to $R^{\mathcal{X}}_t$), while making progressively better choices (as evidenced by the performance with respect to $R'_t$ in column 2 in Figure \ref{fig:regret_plots}).
\end{itemize}

\section{CONCLUSION}
In this work, we have shown how monotonicity of the unknown safety function in a single safety variable allows us to develop a no-regret safe black-box optimization algorithm. We provided several variations of our algorithm that work with different goals of practical relevance. Like many GP-based algorithms, our techniques are primarily suited to low dimensions, and variations suited to higher dimensions would be of interest. Other potential areas of future work include exploring other helpful structures of $g$ beyond monotonicity, and studying extensions to more general settings such as reinforcement learning.

\subsubsection*{Acknowledgement}
This work was supported by the Singapore Ministry of Education Academic Research Fund Tier 1 under grant number A-8000872-00-00.

\newpage
\bibliography{aistats2024}

\begin{thebibliography}{28}
\providecommand{\natexlab}[1]{#1}
\providecommand{\url}[1]{\texttt{#1}}
\expandafter\ifx\csname urlstyle\endcsname\relax
  \providecommand{\doi}[1]{doi: #1}\else
  \providecommand{\doi}{doi: \begingroup \urlstyle{rm}\Url}\fi

\bibitem[Lizotte et~al.(2007)Lizotte, Wang, Bowling, Schuurmans, et~al.]{lizotte2007automatic}
Daniel~J Lizotte, Tao Wang, Michael~H Bowling, Dale Schuurmans, et~al.
\newblock Automatic gait optimization with {G}aussian process regression.
\newblock In \emph{International Joint Conference on Artificial Intelligence}, volume~7, pages 944--949, 2007.

\bibitem[Srinivas et~al.(2012)Srinivas, Krause, Kakade, and Seeger]{srinivas2012information}
Niranjan Srinivas, Andreas Krause, Sham~M Kakade, and Matthias~W Seeger.
\newblock Information-theoretic regret bounds for {G}aussian process optimization in the bandit setting.
\newblock \emph{IEEE Transactions on Information Theory}, 58\penalty0 (5):\penalty0 3250--3265, 2012.

\bibitem[Takahashi and Suzuki(2021)]{takahashi2021mtd}
Ami Takahashi and Taiji Suzuki.
\newblock Bayesian optimization for estimating the maximum tolerated dose in {P}hase {I} clinical trials.
\newblock \emph{Contemporary Clinical Trials Communications}, 21:\penalty0 100753, 2021.

\bibitem[Snoek et~al.(2012)Snoek, Larochelle, and Adams]{snoek2012practical}
Jasper Snoek, Hugo Larochelle, and Ryan~P Adams.
\newblock Practical {B}ayesian optimization of machine learning algorithms.
\newblock \emph{Advances in Neural Information Processing Systems}, 25, 2012.

\bibitem[Vanchinathan et~al.(2014)Vanchinathan, Nikolic, De~Bona, and Krause]{vanchinathan2014explore}
Hastagiri~P Vanchinathan, Isidor Nikolic, Fabio De~Bona, and Andreas Krause.
\newblock Explore-exploit in top-$n$ recommender systems via {G}aussian processes.
\newblock In \emph{ACM Conference on Recommender Systems}, pages 225--232, 2014.

\bibitem[Sui et~al.(2015)Sui, Gotovos, Burdick, and Krause]{sui2015safe}
Yanan Sui, Alkis Gotovos, Joel Burdick, and Andreas Krause.
\newblock Safe exploration for optimization with {G}aussian processes.
\newblock In \emph{International Conference on Machine Learning}, pages 997--1005. PMLR, 2015.

\bibitem[Sui et~al.(2018)Sui, Zhuang, Burdick, and Yue]{sui2018stagewise}
Yanan Sui, Vincent Zhuang, Joel Burdick, and Yisong Yue.
\newblock Stagewise safe {B}ayesian optimization with {G}aussian processes.
\newblock In \emph{International Conference on Machine Learning}, pages 4781--4789. PMLR, 2018.

\bibitem[Berkenkamp et~al.(2021)Berkenkamp, Krause, and Schoellig]{berkenkamp2021bayesian}
Felix Berkenkamp, Andreas Krause, and Angela~P Schoellig.
\newblock Bayesian optimization with safety constraints: Safe and automatic parameter tuning in robotics.
\newblock \emph{Machine Learning}, pages 1--35, 2021.

\bibitem[Berry(2012)]{berry2012adaptive}
Donald~A Berry.
\newblock Adaptive clinical trials in oncology.
\newblock \emph{Nature reviews Clinical oncology}, 9\penalty0 (4):\penalty0 199--207, 2012.

\bibitem[Chevret(2006)]{chevret2006statistical}
S.~Chevret.
\newblock \emph{Statistical Methods for Dose-Finding Experiments}.
\newblock Statistics in Practice. Wiley, 2006.

\bibitem[Cai et~al.(2014)Cai, Yuan, and Ji]{cai2014bayesian}
Chunyan Cai, Ying Yuan, and Yuan Ji.
\newblock A {B}ayesian dose finding design for oncology clinical trials of combinational biological agents.
\newblock \emph{Journal of the Royal Statistical Society Series C: Applied Statistics}, 63\penalty0 (1):\penalty0 159--173, 2014.

\bibitem[Berkenkamp et~al.(2017)Berkenkamp, Turchetta, Schoellig, and Krause]{berkenkamp2017safe}
Felix Berkenkamp, Matteo Turchetta, Angela Schoellig, and Andreas Krause.
\newblock Safe model-based reinforcement learning with stability guarantees.
\newblock \emph{Advances in Neural Information Processing Systems}, 30, 2017.

\bibitem[Turchetta et~al.(2019)Turchetta, Berkenkamp, and Krause]{turchetta2019safe}
Matteo Turchetta, Felix Berkenkamp, and Andreas Krause.
\newblock Safe exploration for interactive machine learning.
\newblock \emph{Advances in Neural Information Processing Systems}, 32, 2019.

\bibitem[Turchetta et~al.(2016)Turchetta, Berkenkamp, and Krause]{turchetta2016safe}
Matteo Turchetta, Felix Berkenkamp, and Andreas Krause.
\newblock Safe exploration in finite {M}arkov decision processes with {G}aussian processes.
\newblock \emph{Advances in Neural Information Processing Systems}, 29, 2016.

\bibitem[Turchetta et~al.(2020)Turchetta, Kolobov, Shah, Krause, and Agarwal]{turchetta2020safe}
Matteo Turchetta, Andrey Kolobov, Shital Shah, Andreas Krause, and Alekh Agarwal.
\newblock Safe reinforcement learning via curriculum induction.
\newblock \emph{Advances in Neural Information Processing Systems}, 33:\penalty0 12151--12162, 2020.

\bibitem[Amani et~al.(2021)Amani, Alizadeh, and Thrampoulidis]{amani2021regret}
Sanae Amani, Mahnoosh Alizadeh, and Christos Thrampoulidis.
\newblock Regret bounds for safe {G}aussian process bandit optimization.
\newblock In \emph{IEEE International Symposium on Information Theory (ISIT)}, pages 527--532, 2021.

\bibitem[Baumann et~al.(2021)Baumann, Marco, Turchetta, and Trimpe]{Baumann2021GoSafeGO}
Dominik Baumann, Alonso Marco, Matteo Turchetta, and Sebastian Trimpe.
\newblock Go{S}afe: Globally optimal safe robot learning.
\newblock \emph{2021 IEEE International Conference on Robotics and Automation (ICRA)}, pages 4452--4458, 2021.

\bibitem[Sukhija et~al.(2022)Sukhija, Turchetta, Lindner, Krause, Trimpe, and Baumann]{Sukhija2022ScalableSE}
Bhavya Sukhija, Matteo Turchetta, David Lindner, Andreas Krause, Sebastian Trimpe, and Dominik Baumann.
\newblock Go{S}afe{O}pt: Scalable safe exploration for global optimization of dynamical systems.
\newblock \emph{Artif. Intell.}, 320:\penalty0 103922, 2022.

\bibitem[Losalka and Scarlett(2023)]{losalka2023benefits}
Arpan Losalka and Jonathan Scarlett.
\newblock Benefits of monotonicity in safe exploration with {G}aussian processes.
\newblock In \emph{Uncertainty in Artificial Intelligence}, pages 1304--1314. PMLR, 2023.

\bibitem[Chowdhury and Gopalan(2017)]{chowdhury2017kernelized}
Sayak~Ray Chowdhury and Aditya Gopalan.
\newblock On kernelized multi-armed bandits.
\newblock In \emph{International Conference on Machine Learning}, pages 844--853. PMLR, 2017.

\bibitem[Vakili et~al.(2021)Vakili, Khezeli, and Picheny]{vakili2021information}
Sattar Vakili, Kia Khezeli, and Victor Picheny.
\newblock On information gain and regret bounds in {G}aussian process bandits.
\newblock In \emph{International Conference on Artificial Intelligence and Statistics}, pages 82--90. PMLR, 2021.

\bibitem[Whitehouse et~al.(2023)Whitehouse, Wu, and Ramdas]{whitehouse2023sublinear}
Justin Whitehouse, Zhiwei~Steven Wu, and Aaditya Ramdas.
\newblock On the sublinear regret of {GP-UCB}.
\newblock \emph{arXiv preprint arXiv:2307.07539}, 2023.

\bibitem[Schreiter et~al.(2015)Schreiter, Nguyen-Tuong, Eberts, Bischoff, Markert, and Toussaint]{schreiter2015safe}
Jens Schreiter, Duy Nguyen-Tuong, Mona Eberts, Bastian Bischoff, Heiner Markert, and Marc Toussaint.
\newblock Safe exploration for active learning with {G}aussian processes.
\newblock In \emph{Joint European Conference on Machine Learning and Knowledge Discovery in Databases}, pages 133--149. Springer, 2015.

\bibitem[Picheny et~al.(2013)Picheny, Wagner, and Ginsbourger]{picheny2013benchmark}
Victor Picheny, Tobias Wagner, and David Ginsbourger.
\newblock A benchmark of kriging-based infill criteria for noisy optimization.
\newblock \emph{Structural and multidisciplinary optimization}, 48:\penalty0 607--626, 2013.

\bibitem[Brockman et~al.(2016)Brockman, Cheung, Pettersson, Schneider, Schulman, Tang, and Zaremba]{brockman2016openai}
Greg Brockman, Vicki Cheung, Ludwig Pettersson, Jonas Schneider, John Schulman, Jie Tang, and Wojciech Zaremba.
\newblock {OpenAI} gym.
\newblock \emph{arXiv preprint arXiv:1606.01540}, 2016.

\bibitem[Hedar(2013)]{hedar2013global}
Abdel-Rahman Hedar.
\newblock {Global optimization test problems}.
\newblock \url{http://www-optima.amp.i.kyoto-u.ac.jp/member/student/hedar/Hedar_files/TestGO.htm}, 2013.
\newblock Accessed: 2024-02-20.

\bibitem[Picheny et~al.(2023)Picheny, Berkeley, Moss, Stojic, Granta, Ober, Artemev, Ghani, Goodall, Paleyes, Vakili, Pascual-Diaz, Markou, Qing, Loka, and Couckuyt]{trieste2023}
Victor Picheny, Joel Berkeley, Henry~B. Moss, Hrvoje Stojic, Uri Granta, Sebastian~W. Ober, Artem Artemev, Khurram Ghani, Alexander Goodall, Andrei Paleyes, Sattar Vakili, Sergio Pascual-Diaz, Stratis Markou, Jixiang Qing, Nasrulloh R. B.~S Loka, and Ivo Couckuyt.
\newblock Trieste: Efficiently exploring the depths of black-box functions with tensorflow, 2023.
\newblock URL \url{https://arxiv.org/abs/2302.08436}.

\bibitem[Shekhar and Javidi(2020)]{shekhar2020multi}
Shubhanshu Shekhar and Tara Javidi.
\newblock Multi-scale zero-order optimization of smooth functions in an {RKHS}.
\newblock \emph{arXiv preprint arXiv:2005.04832}, 2020.

\end{thebibliography}
\bibliographystyle{unsrtnat}

\appendix

\onecolumn

{\centering
    {\huge \bf Appendix \\ [2mm]}  
}

\section{PROOFS}  \label{sec:proofs}

In this section, we present the proofs for the three theorems. 

\subsection{Proof of Theorem 1 (Case 1)}

Let $(s_*, \bx_*)$ denote the optimal action (or any one such action if there are multiple).  Recall the definition of $s^{(\bx)}_t = \max \big\{s \in \dDS : (s,\bx) \in S_t\big\}$ from \eqref{eqn:s_t^x}.

For deriving the following results, we assume the validity of the confidence bounds, which are known to hold with probability at least $1-\delta$.  That is, we condition on \eqref{eq:valid_conf_f} and \eqref{eq:valid_conf_g} both being true. To characterize the regret incurred at a given time instant $t$, we will split the analysis into two cases: (i) the optimal action $(s_*, \bx_*) \notin S_t$, i.e., ${\UCB}_{t-1}^g(s_*, \bx_*) > h$, and (ii) $(s_*, \bx_*) \in S_t$, i.e., ${\UCB}_{t-1}^g(s_*, \bx_*) \leq h$.

\subsubsection{Regret for $(s_*, \bx_*) \notin S_t$}
First, we consider the regret incurred in rounds where $(s_*, \bx_*) \notin S_t$. Since the optimal point $(s_*, \bx_*)$ is safe by definition, we have

\begin{equation}
\label{eqn:g_safe}
    g(s_*, \bx_*) \leq h.
\end{equation}

Next, recall the definition of the set $S_t$ from \eqref{eqn:S_t}, and the definition of $s_t^{(\bx)}$ in \eqref{eqn:s_t^x}. Intuitively, $S_t$ consists of all actions that can be classified as safe via $\UCB^g_{t-1}$ and the safety threshold $h$, while $s_t^{(\bx)}$ corresponds to the action on the current ``safe boundary'' for a specific $\bx$.

The following cases may arise: (i) $\UCB^g_{t-1}(s_t^{(\bx)}, \bx) > h$ (e.g., in the initial rounds when $s_t^{(\bx)}=0$), (ii) $\UCB^g_{t-1}(s_t^{(\bx)}, \bx) = h$ (when $\UCB^g_{t-1}(s,\bx)$ ``crosses'' $h$ for some $s$), or (iii) ${\UCB}^g_{t-1}(s_t^{(\bx)}, \bx) <h$ (for $s_t^{(\bx)}=1$ when $g(1,\bx) < h$). In the following, we first consider the rounds for which either (i) or (ii) holds for $(s_t^{(\bx_t)}, \bx_t)$, i.e,  for the specific $\bx_t$ chosen at round $t$. Thereafter, we show how the results that we derive also continue to remain valid when (iii) holds. 

Since we are assuming that either (i) or (ii) holds for $(s_t^{(\bx_t)}, \bx_t)$ and that $(s_*, \bx_*) \notin S_t$ (for now), we have the following:
\begin{align}
    \UCB_{t-1}^g(s_t^{(\bx_t)}, \bx_t) \geq h, \label{eqn:ucbg_xt_geq_h} \\ 
    \UCB_{t-1}^g(s_t^{(\bx_*)}, \bx_*) \geq h. \label{eqn:ucbg_x*_geq_h}.
\end{align}
In more detail, \eqref{eqn:ucbg_x*_geq_h} follows because the condition $(s_*, \bx_*) \notin S_t$ states that $(s_*, \bx_*)$ has not been discovered as safe by the algorithm, meaning it cannot be that ${\UCB}^g_{t-1}(s_t^{(\bx_*)}, \bx_*) < h$.
\paragraph{Bounding $|\underline{s}_t^{(\bx_t)} - s_t^{(\bx_t)}|$.} 
By the definition of $\underline{s}_t^{(\bx)}$ in \eqref{eqn:us_t^x}, the following holds:
\begin{equation}
\label{eqn:lcb_us}
    \LCB_{t-1}^g(s_t^{(\bx_t)}, \bx_t) + L_g'|\underline{s}_t^{(\bx_t)} - s_t^{(\bx_t)}| \leq h, 
\end{equation}
where strict inequality (i.e., $\LCB_{t-1}^g(s_t^{(\bx_t)}, \bx_t) + L_g'|\underline{s}_t^{(\bx_t)} - s_t^{(\bx_t)}| < h$) may hold when $\underline{s}_t^{(\bx_t)} = 1$.

Therefore, we deduce the following:
\begin{align}
\label{eqn:usxt_sxt_bound}
    L_g'|\underline{s}_t^{(\bx_t)} - s_t^{(\bx_t)}| &\leq h - \LCB_{t-1}^g(s_t^{(\bx_t)}, \bx_t)  \nonumber\\ 
    & \leq \UCB_{t-1}^g(s_t^{(\bx_t)}, \bx_t) - \LCB_{t-1}^g(s_t^{(\bx_t)}, \bx_t) \qquad \text{ (by \eqref{eqn:ucbg_xt_geq_h})}  \nonumber\\ 
    & = 2\beta_t^g \sigma_t^g (s_t^{(\bx_t)}, \bx_t) \qquad \text{ (by \eqref{eq:ucb_f} and \eqref{eq:lcb_f})} \nonumber\\ 
    \implies |\underline{s}_t^{(\bx_t)} - s_t^{(\bx_t)}| & \leq  2\beta_t^g \sigma_t^g (s_t^{(\bx_t)}, \bx_t) / L_g'.
\end{align}
This result upper bounds the distance of $\underline{s}_t^{(\bx_t)}$ from $s_t^{(\bx_t)}$ in terms of the width of the confidence interval at $(s_t^{(\bx_t)}, \bx)$.

Note that \eqref{eqn:usxt_sxt_bound} holds trivially if ${\UCB}^g_{t-1}(s_t^{(\bx_t)}, \bx_t) <h$ (i.e., if \eqref{eqn:ucbg_xt_geq_h} does not hold). This is because the ${\UCB}^g_{t-1}$-value being less than $h$ for an action on the current ``safe boundary'' implies that the algorithm has discovered $(s,\bx_t)$ to be safe for all $s \in \dDS$, and hence $s_t^{(\bx_t)} = 1$. This would also imply that $\usxt = 1$ (following \eqref{eqn:us_t^x}), thus giving $|\underline{s}_t^{(\bx_t)} - s_t^{(\bx_t)}| = 0$.

\paragraph{Bounding $|s_* - s_t^{(\bx_*)}|$.} Similarly to the above, due to the safety of the optimal action, we have the following:
\begin{align}
\label{eqn:s*_stx*_bound}
    L_g'|s_* - s_t^{(\bx_*)}| &\leq g(s_*, \bx_*) - g(s_t^{(\bx_*)}, \bx_*)  \qquad \text{(by definition of $L_g'$)} \nonumber\\
    & \leq h - \LCB_{t-1}^g(s_t^{(\bx_*)}, \bx_*)  \qquad \text{ (by \eqref{eq:valid_conf_g} and \eqref{eqn:g_safe})}  \nonumber\\ 
    &\leq \UCB_{t-1}^g(s_t^{(\bx_*)}, \bx_*) - \LCB_{t-1}^g(s_t^{(\bx_*)}, \bx_*)  \qquad \text{ (by \eqref{eqn:ucbg_x*_geq_h})} \nonumber\\ 
    &= 2\beta_t^g \sigma_{t-1}^g(s_t^{(\bx_*)}, \bx_*) \qquad \text{ (by \eqref{eq:ucb_f} and \eqref{eq:lcb_f})}\nonumber\\
    \implies |s_* - s_t^{(\bx_*)}| &\leq 2\beta_t^g \sigma_{t-1}^g(s_t^{(\bx_*)}, \bx_*)/L_g'.
\end{align}
This result upper bounds the distance of $s_*$ from $s_t^{(\bx_*)}$ in terms of the width of the confidence interval at $(s_t^{(\bx_*)}, \bx_*)$.

\paragraph{Bounding instantaneous regret.}
We consider the instantaneous regret $f(s_*, \bx_*) - f(s_t, \bx_t)$ in each round $t$, and split it as a sum of three terms, which we proceed to bound individually:
\begin{equation}
    f(s_*, \bx_*) - f(s_t, \bx_t) = \left(f(s_*,\bx_*) - f(s_t^{(\bx_*)}, \bx_*)\right) + \left(f(s_t^{(\bx_*)}, \bx_*) - f(s_t^{(\bx_t)}, \bx_t) \right) + \left( f(s_t^{(\bx_t)}, \bx_t) - f(s_t, \bx_t)\right). \label{eq:regret_split}
\end{equation}
In some cases, we will also use a near-identical decomposition with $\hat{s}_t^{(\bx_t)}$ (from \eqref{eqn:hat_s_t^x}) replacing $s_t^{(\bx_t)}$ (depending on the criteria for $\bx_t$ not being eliminated at round $t$) as follows:
\begin{equation}
    f(s_*, \bx_*) - f(s_t, \bx_t) = \left(f(s_*,\bx_*) - f(s_t^{(\bx_*)}, \bx_*)\right) + \left(f(s_t^{(\bx_*)}, \bx_*) - f(\hat{s}_t^{(\bx_t)}, \bx_t) \right) + \left( f(\hat{s}_t^{(\bx_t)}, \bx_t) - f(s_t, \bx_t)\right). \label{eq:regret_split_ii}
\end{equation}

We also note that given the acquisition function in \eqref{eqn:acq_i}, $(s_t, \bx_t)$ being chosen at round $t$ implies that
\begin{equation}
\label{eqn:max_beta_sigma}
    \max\left\{\beta_t^g \sigma_{t-1}^g(s_t, \bx_t), \beta_t^f \sigma_{t-1}^f(s_t, \bx_t)\right\} \geq \max \left\{ \max_{(s,\bx) \in G_t}\Big\{\beta^g_t\sigma^g_{t-1}(s,\bx)\Big\}, \max_{(s,\bx) \in G_t \cup M_t}\left\{\beta^f_t\sigma^f_{t-1}(s,\bx)\right\} \right\}.
\end{equation}

We bound the first term in \eqref{eq:regret_split} as follows:
\begin{align}
    f(s_*,\bx_*) - f(s_t^{(\bx_*)}, \bx_*) & \leq L_f|s_* - s_t^{(\bx_*)}|  \qquad \text{ (by definition of $L_f$)} \nonumber\\ 
    & \leq 2(L_f/L_g')\beta_t^g \sigma_{t-1}^g(s_t^{(\bx_*)}, \bx_*)  \qquad \text{ (by \eqref{eqn:s*_stx*_bound})} \nonumber\\ 
    & \leq 2(L_f/L_g')\cdot\max\{\beta_t^g \sigma_{t-1}^g(s_t, \bx_t), \beta_t^f \sigma_{t-1}^f(s_t, \bx_t)\}  \qquad \text{ (by \eqref{eqn:max_beta_sigma})}.
    \label{eqn:f_bound_1}
\end{align}

We now consider the second term in \eqref{eq:regret_split}. Since $\bx_t$ was chosen instead of $\bx_*$, we have that $\bx_t$ was not eliminated. This implies that at least one of the two elimination criteria did not hold (note that both \eqref{eqn:elim_i_i} and \eqref{eqn:elim_i_ii} must hold for $\bx_t$ to be eliminated). If \eqref{eqn:elim_i_i} did not hold, then we have 
\begin{align}
    f(s_t^{(\bx_*)}, \bx_*) - f(\hat{s}_t^{(\bx_t)}, \bx_t) &\leq \UCB^f_{t-1}(s_t^{(\bx_*)}, \bx_*) - \LCB^f_{t-1}(\hat{s}_t^{(\bx_t)}, \bx_t) \nonumber \\
    & = \LCB^f_{t-1}(s_t^{(\bx_*)}, \bx_*) + 2\beta_t^f \sigma_{t-1}^f(s_t^{(\bx_*)}, \bx_*) \nonumber \\ 
    & \qquad \qquad - \UCB^f_{t-1}(\hat{s}_t^{(\bx_t)}, \bx_t) +  2\beta_t^f \sigma_{t-1}^f(\hat{s}_t^{(\bx_t)}, \bx_t) \nonumber\\
    & \leq 2\beta_t^f \sigma_{t-1}^f(s_t^{(\bx_*)}, \bx_*) +  2\beta_t^f \sigma_{t-1}^f(\hat{s}_t^{(\bx_t)}, \bx_t)  \qquad \text{ (by \eqref{eqn:elim_i_i} being \texttt{false})}\\
    & \leq 4\cdot \max\{\beta_t^g \sigma_{t-1}^g(s_t, \bx_t), \beta_t^f \sigma_{t-1}^f(s_t, \bx_t)\}   \qquad \text{ (by \eqref{eqn:max_beta_sigma})} \label{eqn:f_bound_2}
\end{align}
Alternatively, if \eqref{eqn:elim_i_ii} did not hold, then we have
\begin{align}
    f(s_t^{(\bx_*)}, \bx_*) & - f(s_t^{(\bx_t)}, \bx_t) \leq \UCB^f_{t-1}(s_t^{(\bx_*)}, \bx_*) - \LCB^f_{t-1}(s_t^{(\bx_t)}, \bx_t) \nonumber \\
    & = \LCB^f_{t-1}(s_t^{(\bx_*)}, \bx_*) + 2\beta_t^f \sigma_{t-1}^f(s_t^{(\bx_*)}, \bx_*) \nonumber \\ 
    & \qquad \qquad - \UCB^f_{t-1}(s_t^{(\bx_t)}, \bx_t) +  2\beta_t^f \sigma_{t-1}^f(s_t^{(\bx_t)}, \bx_t) \nonumber\\
    & \leq 2\beta_t^f \sigma_{t-1}^f(s_t^{(\bx_*)}, \bx_*) +  2\beta_t^f \sigma_{t-1}^f(s_t^{(\bx_t)}, \bx_t) + L_f|\underline{s}^{(\bx_t)}_t - s^{(\bx_t)}_t|  \qquad \text{ (by \eqref{eqn:elim_i_ii} being \texttt{false})}\\ 
    & \leq 2\beta_t^f \sigma_{t-1}^f(s_t^{(\bx_*)}, \bx_*) +  2\beta_t^f \sigma_{t-1}^f(s_t^{(\bx_t)}, \bx_t) + 2(L_f/L_g')\beta_t^g \sigma_t^g (s_t^{(\bx_t)}, \bx_t)  \qquad \text{ (by \eqref{eqn:usxt_sxt_bound})}\\
    & \leq 4 \cdot\max\left\{\beta_t^g \sigma_{t-1}^g(s_t, \bx_t), \beta_t^f \sigma_{t-1}^f(s_t, \bx_t)\right\}  \\
    & \qquad \qquad \qquad + 2(L_f/L_g')\max\left\{\beta_t^g \sigma_{t-1}^g(s_t, \bx_t), \beta_t^f \sigma_{t-1}^f(s_t, \bx_t)\right\}  \qquad \text{ (by \eqref{eqn:max_beta_sigma}).}
\end{align}

Finally, we consider the third term in \eqref{eq:regret_split}. Suppose again that $\bx_t$ remained non-eliminated due to \eqref{eqn:elim_i_i} being \texttt{false}. In this case, we have:
\begin{align}
    f(s_t^{(\bx_t)}, \bx_t) - f(s_t, & \bx_t) \leq \begin{cases}
            0, & \text{(if $s_t = s_t^{(\bx_t)}$)}\\
            \UCB^f_{t-1}(s_t^{(\bx_t)}, \bx_t) - \LCB^f_{t-1}(\hat{s}_t^{(\bx_t)}, \bx_t), & \text{(if $s_t = \hat{s}_t^{(\bx_t)}$)}
            \end{cases} \nonumber \\ 
    & \leq \max\{0, \UCB^f_{t-1}(s_t^{(\bx_t)}, \bx_t) - \UCB^f_{t-1}(\hat{s}_t^{(\bx_t)}, \bx_t) + 2\beta_t^f\sigma_{t-1}^f(\hat{s}_t^{(\bx_t)}, \bx_t)\} \nonumber \\
    & \leq 2\beta_t^f\sigma_{t-1}^f(\hat{s}_t^{(\bx_t)}, \bx_t) \label{eq:explain_i} \\
    & \leq 2\cdot\max\{\beta_t^g \sigma_{t-1}^g(s_t, \bx_t), \beta_t^f \sigma_{t-1}^f(s_t, \bx_t)\}   \qquad \text{ (by \eqref{eqn:max_beta_sigma}),}
\end{align}
where \eqref{eq:explain_i} holds because $\UCB^f_{t-1}(\hat{s}_t^{(\bx_t)}, \bx_t) \ge  \UCB^f_{t-1}(s_t^{(\bx_t)}, \bx_t)$ by the definition of $\hat{s}_t^{(\bx_t)}$ in \eqref{eqn:hat_s_t^x}.
Regarding the first step above, we note that either $s_t = s_t^{(\bx_t)}$ or $s_t = \hat{s}_t^{(\bx_t)}$ must hold, because the actions chosen in any round must belong to either $G_t$ or $M_t$.

Next, if $\bx_t$ remained non-eliminated due to \eqref{eqn:elim_i_ii} being \texttt{false}, then we can bound the third term in \eqref{eq:regret_split_ii} as follows:
\begin{align}
    f(\hat{s}_t^{(\bx_t)}, & \bx_t) - f(s_t, \bx_t) \leq \begin{cases}
            0, & \text{(if $s_t = \hat{s}_t^{(\bx_t)}$)}\\
            \UCB^f_{t-1}(\hat{s}_t^{(\bx_t)}, \bx_t) - \LCB^f_{t-1}(s_t^{(\bx_t)}, \bx_t), & \text{(if $s_t = s_t^{(\bx_t)}$)}
            \end{cases} \nonumber \\ 
    & \leq \max\{0, \LCB^f_{t-1}(\hat{s}_t^{(\bx_t)}, \bx_t) - \UCB^f_{t-1}(s_t^{(\bx_t)}, \bx_t) \nonumber \\ 
    & \qquad \qquad \qquad + 2\beta_t^f\sigma_{t-1}^f(s_t^{(\bx_t)}, \bx_t) + 2\beta_t^f\sigma_{t-1}^f(s_t^{(\bx_t)}, \bx_t)\} \nonumber \\
    & \leq \max\{0, L_f|\underline{s}_t^{(\bx_t)} - s_t^{(\bx_t)}| + 2\beta_t^f\sigma_{t-1}^f(s_t^{(\bx_t)}, \bx_t) + 2\beta_t^f\sigma_{t-1}^f(s_t^{(\bx_t)}, \bx_t)\}   \qquad \text{ (by \eqref{eqn:elim_i_ii} being \texttt{false})}\nonumber \\ 
    &\leq 2(L_f/L_g')\beta_t^g\sigma_{t-1}^g(s_t^{(\bx_t)}, \bx_t) +  2\beta_t^f\sigma_{t-1}^f(s_t^{(\bx_t)}, \bx_t) + 2\beta_t^f\sigma_{t-1}^f(s_t^{(\bx_t)}, \bx_t)  \quad~ \text{ (by \eqref{eqn:usxt_sxt_bound}}) \\
    & \leq 2(L_f/L_g')\max\{\beta_t^g \sigma_{t-1}^g(s_t, \bx_t), \beta_t^f \sigma_{t-1}^f(s_t, \bx_t)\} \\
    & \qquad \qquad \qquad + 4\cdot\max\{\beta_t^g \sigma_{t-1}^g(s_t, \bx_t), \beta_t^f \sigma_{t-1}^f(s_t, \bx_t)\}   \qquad \text{ (by \eqref{eqn:max_beta_sigma}).}
\end{align}

Combining the above three terms, we obtain the following bound on the \textit{instantaneous regret} incurred by the algorithm (irrespective of whether $(s_t,\bx_t) \in G_t$ or $(s_t,\bx_t) \in M_t$, and irrespective of the condition that caused $\bx_t$ to remain non-eliminated):
\begin{equation}
\label{eqn:inst_regret_i}
    f(s_*, \bx_*) - f(s_t, \bx_t) \leq \left(8 + \frac{6L_f}{L_g'}\right) \max\{\beta_t^g \sigma_{t-1}^g(s_t, \bx_t), \beta_t^f \sigma_{t-1}^f(s_t, \bx_t)\}.
\end{equation}

\subsubsection{Regret for $(s_*, \bx_*) \in S_t$}
Next, we consider $(s_*, \bx_*) \in S_t$, i.e., ${\UCB}^g_{t-1}(s_*, \bx_*) \le h$. In this case, it must hold that $s_* \leq s^{(\bx_*)}_t$ (since $s^{(\bx_*)}_t$ is defined as the ``highest'' safe $s$ for $\bx_*$ in \eqref{eqn:s_t^x}). Thus, we have
\begin{equation}
    \UCB_{t-1}^f(s_*, \bx_*) \leq \UCB_{t-1}^f(\hat{s}_t^{(\bx_*)}, \bx_*), \label{eq:ucb*}
\end{equation} 
since $\hat{s}_t^{(\bx_*)}$ is the maximizer of $\UCB_{t-1}^f(\cdot, \bx_*)$ over $s \leq s^{(\bx_*)}_t$. In this case, we can bound the instantaneous regret \eqref{eq:regret_split} directly as follows:
\begin{align}
    f(s_*, \bx_*) - f(s_t, & \bx_t) \leq \UCB_{t-1}^f(s_*, \bx_*) - \LCB_{t-1}^f(s_t, \bx_t) \nonumber \\ 
    & \leq \UCB_{t-1}^f(\hat{s}_t^{(\bx_*)}, \bx_*) - \UCB_{t-1}^f(s_t, \bx_t) + 2\beta_t^f\sigma_{t-1}^f(s_t, \bx_t)  \qquad \text{ (by \eqref{eq:ucb*}} \nonumber \\ 
    & = \LCB_{t-1}^f(\hat{s}_t^{(\bx_*)}, \bx_*) - \UCB_{t-1}^f(s_t, \bx_t) \nonumber \\
    & \qquad \qquad \qquad + 2\beta_t^f\sigma_{t-1}^f(s_t, \bx_t) + 2\beta_t^f\sigma_{t-1}^f(\hat{s}_t^{(\bx_*)}, \bx_*) \nonumber \qquad \text{ (by \eqref{eq:ucb_f})} \\ 
    & \leq L_f|\underline{s}_t^{(\bx_t)} - s_t^{(\bx_t)}| + 2\beta_t^f\sigma_{t-1}^f(s_t, \bx_t) + 2\beta_t^f\sigma_{t-1}^f(\hat{s}_t^{(\bx_*)}, \bx_*) \label{eqn:to_explain} \qquad \text{ (see \eqref{eq:explation_ii} below)}\\
    & \leq 2(L_f/L_g')\beta_t^g\sigma_{t-1}^g(s_t^{(\bx_t)}, \bx_t) + 2\beta_t^f\sigma_{t-1}^f(s_t, \bx_t) + 2\beta_t^f\sigma_{t-1}^f(\hat{s}_t^{(\bx_*)}, \bx_*)   \qquad \text{ (by \eqref{eqn:usxt_sxt_bound})} \nonumber \\
    & \leq 2(L_f/L_g')\max\left\{\beta_t^g \sigma_{t-1}^g(s_t, \bx_t), \beta_t^f \sigma_{t-1}^f(s_t, \bx_t)\right\} \nonumber\\
    & \qquad \qquad \qquad + 4\cdot\max\left\{\beta_t^g \sigma_{t-1}^g(s_t, \bx_t), \beta_t^f \sigma_{t-1}^f(s_t, \bx_t)\right\}    \qquad \text{ (by \eqref{eqn:max_beta_sigma}),} \label{eqn:inst_regret_ii}
\end{align}
where \eqref{eqn:to_explain} holds because of the following:
\begin{align}
     \LCB_{t-1}^f(\hat{s}_t^{(\bx_*)}, \bx_*) & - \UCB_{t-1}^f(s_t, \bx_t) \nonumber \\
     &  \leq \begin{cases}
            \max\left\{0, L_f|\underline{s}_t^{(\bx_t)} - s_t^{(\bx_t)}|\right\} \text{, if } s_t = \hat{s}_t^{(\bx_t)} \quad \text{ (since $\mathtt{elim}_t(\bx_t)$  = \texttt{false}))} \\ 
            L_f|\underline{s}_t^{(\bx_t)} - s_t^{(\bx_t)}| \text{, if }s_t = s_t^{(\bx_t)} \quad \text{ (since $\mathtt{expd}_t(\bx_t)$  = \texttt{true}) }.
     \end{cases} \label{eq:explation_ii}
\end{align}
Here, we again use the fact that either $s_t = \hat{s}_t^{(\bx_t)}$ or $s_t = {s}_t^{(\bx_t)}$ (as the action $(s_t, \bx_t)$ chosen at round $t$ must belong to either $M_t$ or $G_t$). Therefore, either \eqref{eqn:elim_i_i} or \eqref{eqn:elim_i_ii} must be \texttt{false} when $s_t = \hat{s}_t^{(\bx_t)}$ (since $\bx_t$ was not eliminated), and \eqref{eqn:exp_i} must be \texttt{true} when $s_t = {s}_t^{(\bx_t)}$ (since an action on the safe boundary is only chosen if it considered ``potentially beneficial'' to expand, as decided by $\mathtt{expd}_t$). 

\subsubsection{Bounding the cumulative regret}
Summing the instantaneous regret terms in \eqref{eqn:inst_regret_i} and \eqref{eqn:inst_regret_ii} from $t=1, \dots, T$, we get the following:
\begin{align}
    R_T = \sum_{t=1}^T \left(f(s_*, \bx_*) - f(s_t, \bx_t) \right) \, 
    & \leq \left(8 + \frac{6L_f}{L_g'}\right) \sum_{t=1}^T \max\left\{\beta_t^g \sigma_{t-1}^g(s_t, \bx_t), \beta_t^f \sigma_{t-1}^f(s_t, \bx_t)\right\} \\
    & \leq \left(8 + \frac{6L_f}{L_g'}\right) \sum_{t=1}^T \left(\beta_t^g \sigma_{t-1}^g(s_t, \bx_t) + \beta_t^f \sigma_{t-1}^f(s_t, \bx_t)\right) \label{eqn:mod_acq_ref_i} \\ 
    & \leq \left(8 + \frac{6L_f}{L_g'}\right)\left( \beta_T^g\sum_{t=1}^T \sigma_{t-1}^g(s_t, \bx_t) + \beta_T^f \sum_{t=1}^T \sigma_{t-1}^f(s_t, \bx_t)\right),
\end{align}
where we used the assumption that $\beta_t^f, \beta_t^g$ are non-decreasing with respect to $t$. Then, from Lemma 4 of \citep{chowdhury2017kernelized}, we have the standard bounds:
\begin{align}
\sum_{t=1}^{T} \sigma^g_{t-1}\left(s_t, \bx_{t}\right) = O\big(\sqrt{T\gamma^g_{T}}\big),   \\ 
\sum_{t=1}^{T} \sigma^f_{t-1}\left(s_t, \bx_{t}\right) = O\big(\sqrt{T\gamma^f_{T}}\big).
\end{align}

Hence, with probability at least $1-\delta$,
\begin{equation}
    R_T = O\left( \left(1+\frac{L_f}{L'_g}\right) \left(\beta^g_T\sqrt{T\gamma^g_T} + \beta^f_T\sqrt{T\gamma^f_T} \right) \right).
\end{equation}
Specifically, for the choice of $\beta_f$ and $\beta_g$ from Lemma \ref{lemma:beta}, we have the following:
\begin{align}
R_{T} = O\biggl( B_g \left(1+\frac{L_f}{L'_g}\right)& \sqrt{T \gamma^g_{T}}+R_g \left(1+\frac{L_f}{L'_g}\right)\sqrt{T \gamma^g_{T}\left(\gamma^g_{T}+\ln (1 / \delta)\right)} \nonumber \\ 
& + B_f \left(1+\frac{L_f}{L'_g}\right) \sqrt{T \gamma^f_{T}}+R_f \left(1+\frac{L_f}{L'_g}\right)\sqrt{T \gamma^f_{T}\left(\gamma^f_{T}+\ln (1 / \delta)\right)},
\end{align}
where $\beta^g_T \leq B_g + R_g \sqrt{2\left(\gamma^g_{T}+1+\ln (2 / \delta)\right)}$ since $\gamma^g_t$ is monotonically increasing (and similarly for $\beta^f_T$).

\subsection{Proof of Theorem 1 (Case 3)}

This case turns out to be a minor variation of the above, so we omit the full details and only describe the differences.  
First, we note that the initial results derived with respect to the safety function $g$ hold in this case as well. Specifically, \eqref{eqn:g_safe} to \eqref{eqn:s*_stx*_bound} are valid because of the same arguments as presented in the previous proof. 

Next, \eqref{eqn:max_beta_sigma} also holds given the modified acquisition function for case 3, with the only change being that the set $M_t$ is not considered here. Also, note that the optimal safe action is guaranteed to lie on the ``safe boundary'' in this case. We first consider the scenario where $(s_*, \bx_*) \notin S_t$, which implies that \eqref{eqn:f_bound_1} holds without modification.

Since the elimination criteria is simplified, following \eqref{eqn:f_bound_2}, we now have
\begin{equation}
    \label{eqn:f_bound_2_3}
    f(s_t^{(\bx_*)}, \bx_*) - f(s_t^{(\bx_t)}, \bx_t) \leq  \left(4 + \frac{2L_f}{L_g'}\right)\max\left\{\beta_t^g \sigma_{t-1}^g(s_t, \bx_t), \beta_t^f \sigma_{t-1}^f(s_t, \bx_t)\right\}.
\end{equation}

Finally, since $s_t = s_t^{(\bx_t)}$ for all $t \geq 1$ in this case, we can combine \eqref{eqn:f_bound_1} and \eqref{eqn:f_bound_2_3} to obtain a bound on the instantaneous regret as follows:
\begin{equation}
    f(s_*, \bx_*) - f(s_t, \bx_t) \leq \left(4 + \frac{4L_f}{L_g'}\right) \max\left\{\beta_t^g \sigma_{t-1}^g(s_t, \bx_t), \beta_t^f \sigma_{t-1}^f(s_t, \bx_t)\right\}.
    \label{eqn:case_3_regret_i}
\end{equation}
It may also hold that $(s_*, \bx_*) \in S_t$ when $s_* = 1$. In this case, we bound the instantaneous regret directly as follows:
\begin{align}
    f(s_*, \bx_*) - f(s_t, & \bx_t) \leq \UCB_{t-1}^f(s_*, \bx_*) - \LCB_{t-1}^f(s_t, \bx_t) \nonumber \\ 
    & = \UCB_{t-1}^f({s}_*, \bx_*) - \UCB_{t-1}^f(s_t, \bx_t) + 2\beta_t^f\sigma_{t-1}^f(s_t, \bx_t)  \nonumber \\ 
    & = \LCB_{t-1}^f({s}_*, \bx_*) - \UCB_{t-1}^f(s_t, \bx_t) + 2\beta_t^f\sigma_{t-1}^f(s_t, \bx_t) + 2\beta_t^f\sigma_{t-1}^f({s}_*, \bx_*) \nonumber  \\ 
    & \leq 2\beta_t^f\sigma_{t-1}^f(s_t, \bx_t) + 2\beta_t^f\sigma_{t-1}^f({s}_*, \bx_*) \qquad \text{ (by \eqref{eqn:elim_i_ii} being false, and since $s_t^{(\bx_*)} = \underline{s}_t^{(\bx_*)}$)} \nonumber \\
    & \leq 4\cdot\max\left\{\beta_t^g \sigma_{t-1}^g(s_t, \bx_t), \beta_t^f \sigma_{t-1}^f(s_t, \bx_t)\right\}    \qquad \text{ (by \eqref{eqn:max_beta_sigma}).} 
    \label{eqn:case_3_regret_ii}
\end{align}
The cumulative regret bound follows similarly to the previous proof by summing over the instantaneous regret terms in \eqref{eqn:case_3_regret_i} and \eqref{eqn:case_3_regret_ii}. 

\subsection{Proof of Theorem 2}

Similar to the proof of Theorem \ref{theorem:regret_bound}, we split the proof into two possible scenarios based on the chosen $\bx_t$ in each round: (i) $(s^{(\bx_t)}_*, \bx_t) \notin S_t$, i.e., ${\UCB}_{t-1}^g(s^{(\bx_t)}_*, \bx_t) > h$  and (ii) $(s^{(\bx_t)}_*, \bx_t) \in S_t$, i.e., ${\UCB}_{t-1}^g(s^{(\bx_t)}_*, \bx_t) \leq h$.

\subsubsection{Regret for $(s^{(\bx_t)}_*, \bx_t) \notin S_t$}
Following the proof of Theorem \ref{theorem:regret_bound}, several results concerning the safety function $g$ hold. Specifically, the inequalities not concerning $\bx_*$ from \eqref{eqn:ucbg_xt_geq_h} to \eqref{eqn:usxt_sxt_bound} continue to hold for the modified algorithm, due to the same arguments as presented earlier (with \eqref{eqn:ucbg_x*_geq_h} skipped).  However, for finding the regret $r_t'$ here, we now consider the optimal $s$ corresponding to $\bx_t$ (i.e., $s^{(\bx_t)}_*$) as defined in \eqref{eqn:cum_regret_ii}, instead of the global safe optimum action. The inequalities derived earlier are modified accordingly as follows. 

Corresponding to \eqref{eqn:g_safe}, we now have
\begin{equation}
\label{eqn:g_safe_ii}
    g(s^{(\bx_t)}_*, \bx_t) \leq h.
\end{equation}

Moreover, from \eqref{eqn:g_safe_ii} and \eqref{eqn:ucbg_xt_geq_h}, we have:
\begin{align}
\label{eqn:s*_stx*_bound_ii}
    L_g'|s^{(\bx_t)}_* - s_t^{(\bx_t)}| & \leq g(s^{(\bx_t)}_*, \bx_t) - g(s_t^{(\bx_t)}, \bx_t)   \qquad \text{ (by the definition of $L_g'$)} \nonumber\\
    & \leq h - \LCB_{t-1}^g(s_t^{(\bx_t)}, \bx_t) \qquad \text{ (by \eqref{eqn:g_safe_ii})} \nonumber\\ 
    &\leq \UCB_{t-1}^g(s_t^{(\bx_t)}, \bx_t) - \LCB_{t-1}^g(s_t^{(\bx_t)}, \bx_t) \qquad \text{ (by \eqref{eqn:ucbg_xt_geq_h}} \nonumber\\ 
    &= 2\beta_t^g \sigma_{t-1}^g(s_t^{(\bx_t)}, \bx_t) \nonumber\\
    \implies |s^{(\bx_t)}_* - s_t^{(\bx_t)}| &\leq 2\beta_t^g \sigma_{t-1}^g(s_t^{(\bx_t)}, \bx_t)/L_g'. 
\end{align}
Next, with respect to the objective function $f$, we have
\begin{align}
    f(s^{(\bx_t)}_*, \bx_t) - f(s^{(\bx_t)}_t, \bx_t) & \leq L_f|s^{(\bx_t)}_* - s^{(\bx_t)}_t| \qquad \text{ (by definition of $L_f$)}\nonumber \\ 
    &\leq 2(L_f/L_g')\beta_t^g \sigma_{t-1}^g(s_t^{(\bx_t)}, \bx_t) \qquad \text{ (by \eqref{eqn:s*_stx*_bound_ii}).} \label{eq:add_i}
\end{align}
In addition, we have the following:
\begin{align}
    f(s^{(\bx_t)}_t, \bx_t) - f(s_t, \bx_t) &\leq \begin{cases}
        0, & \text{(if } s_t = s^{(\bx_t)}_t) \\
        \UCB^f_{t-1}(s^{(\bx_t)}_t, \bx_t) - \LCB^f_{t-1}(\hat{s}^{(\bx_t)}_t, \bx_t), & \text{(if } s_t = \hat{s}^{(\bx_t)}_t \text{, by \eqref{eq:valid_conf_f})}
    \end{cases} \\ 
    & \leq \UCB^f_{t-1}(\hat{s}^{(\bx_t)}_t, \bx_t) - \LCB^f_{t-1}(\hat{s}^{(\bx_t)}_t, \bx_t) \\
    & \leq 2\beta^f_t\sigma^f_{t-1}(\hat{s}^{(\bx_t)}_t, \bx_t), \label{eq:add_ii}
\end{align}
where we again use the fact that the chosen action must belong to either $G_t$ (i.e, $s_t = s^{(\bx_t)}_t$) or $M_t$ (i.e., $s_t = \hat{s}^{(\bx_t)}_t$).

Adding \eqref{eq:add_i} and \eqref{eq:add_ii}, we obtain
\begin{align}
     f(s^{(\bx_t)}_*, \bx_t) -  f(s_t, \bx_t) & \leq  2(L_f/L_g')\beta_t^g \sigma_{t-1}^g(s_t^{(\bx_t)}, \bx_t) + 2\beta^f_t\sigma^f_{t-1}(\hat{s}^{(\bx_t)}_t, \bx_t) \nonumber \\ 
     & \leq \left(2 + \frac{2L_f}{L_g'}\right)\max\{\beta_t^g \sigma_{t-1}^g(s_t, \bx_t), \beta_t^f \sigma_{t-1}^f(s_t, \bx_t)\}.
\end{align}

\subsubsection{Regret for $(s^{(\bx_t)}_*, \bx_t) \in S_t$}
When $(s^{(\bx_t)}_*, \bx_t) \in S_t$, similarly to earlier as in \eqref{eq:ucb*}, we have:
\begin{equation}
    \UCB^f_{t-1}(s^{(\bx_t)}_*, \bx_t) \leq \UCB^f_{t-1}(\hat{s}^{(\bx_t)}, \bx_t) \label{eq:ucb*_2}.
\end{equation}

In this case, we can bound the instantaneous regret $r_t' = f(s^{(\bx_t)}_*, \bx_t) -  f(s_t, \bx_t)$ directly as follows:
\begin{align}
    f(s^{(\bx_t)}_*, \bx_t) -  f(s_t, \bx_t) & \leq \UCB^f_{t-1}(s^{(\bx_t)}_*, \bx_t) -  \LCB^f_{t-1}(s_t, \bx_t) \qquad \text{ (by \eqref{eq:valid_conf_f})} \nonumber \\ 
    & \leq \begin{cases}
        2\beta^f_t\sigma^f_{t-1}(\hat{s}^{(\bx_t)}, \bx_t) \text{, if } s_t = \hat{s}^{(\bx_t)}_t, \qquad (\text{by \eqref{eq:ucb*_2}} )\\
        \UCB^f_{t-1}(s^{(\bx_t)}_*, \bx_t) -  \LCB^f_{t-1}(s^{(\bx_t)}_t, \bx_t) \text{, if }s_t = s^{(\bx_t)}_t. 
    \end{cases} \label{eqn:f_bound_2_2}
\end{align}

Furthermore, we have the following when $s_t = s_t^{(\bx_t)}$:
\begin{align}
    \UCB^f_{t-1}(s^{(\bx_t)}_*, & \bx_t) -  \LCB^f_{t-1}(s^{(\bx_t)}_t, \bx_t) \leq  
    \LCB^f_{t-1}(s^{(\bx_t)}_*, \bx_t) -  \UCB^f_{t-1}(s^{(\bx_t)}_t, \bx_t) \\
    & \qquad \qquad \qquad \qquad \qquad \qquad 2\beta^f_t\sigma^f_{t-1}(s^{(\bx_t)}_*, \bx_t) + 2\beta^f_t\sigma^f_{t-1}(s^{(\bx_t)}_t, \bx_t) \\ 
    & \leq L_f|\underline{s}_t^{(\bx_t)} - s_t^{(\bx_t)}| + 2\beta^f_t\sigma^f_{t-1}(s^{(\bx_t)}_*, \bx_t) + 2\beta^f_t\sigma^f_{t-1}(s^{(\bx_t)}_t, \bx_t) \qquad \text{ (by \eqref{eqn:exp_ii}})\\ 
    & \leq 2(L_f/L_g')\beta^g_t\sigma^g_{t-1}(s^{(\bx_t)}_t, \bx_t) + 2\beta^f_t\sigma^f_{t-1}(s^{(\bx_t)}_*, \bx_t) + 2\beta^f_t\sigma^f_{t-1}(s^{(\bx_t)}_t, \bx_t) \qquad \text{ (by \eqref{eqn:usxt_sxt_bound}}) \\ 
    &\leq 2\left((L_f/L_g') +  2\right) \max\{\beta_t^g \sigma_{t-1}^g(s_t, \bx_t), \beta_t^f \sigma_{t-1}^f(s_t, \bx_t)\}  \qquad \text{ (by \eqref{eqn:max_beta_sigma}}).
\end{align}
Combining this with \eqref{eqn:f_bound_2_2}, we obtain
\begin{equation}
    f(s^{(\bx_t)}_*, \bx_t) -  f(s_t, \bx_t) \leq 2\left((L_f/L_g') +  2\right) \max\{\beta_t^g \sigma_{t-1}^g(s_t, \bx_t), \beta_t^f \sigma_{t-1}^f(s_t, \bx_t)\}.
\end{equation}

Finally, combining the results from $t=1, \dots, T$,
we can conclude that
\begin{align}
    R_T' = \sum_{t=1}^T r_t' & = \sum_{t=1}^T \left(f(s^{(\bx_t)}_*, \bx_t) -  f(s_t, \bx_t) \right) \\ 
    & \leq 2\left((L_f/L_g') +  2\right) \max\left\{\beta_t^g \sigma_{t-1}^g(s_t, \bx_t), \beta_t^f \sigma_{t-1}^f(s_t, \bx_t)\right\}.
\end{align}
As before, we can use Lemma 4 from \citep{chowdhury2017kernelized} to conclude that with probability at least $1-\delta$,
\begin{equation}
R_T' = O\left( \left(1+\frac{L_f}{L'_g}\right) \left(\beta^g_T\sqrt{T\gamma^g_T} + \beta^f_T\sqrt{T\gamma^f_T} \right) \right). \tag*{\qedsymbol}
\end{equation}

\subsection{Proof of Theorem 3}

All results concerning the safety function $g$ from the previous theorem's proof hold here. However, when considering the objective function $f$, note that we are no longer concerned with the instantaneous regret incurred with respect to the action chosen by the algorithm. Instead, we now consider the best estimate $\hat{s}_t^{(\bx)}$ for any given $\bx$ at round $t$, and try to bound the worst-case (over $\bx \in \dDX$) simple regret incurred by the algorithm.

Suppose that the worst-case simple regret is incurred by $\bx = \underline{\bx}_t$ at round $t$, i.e., 
\begin{equation}
    \underline{\bx}_t = \argmax_{\bx \in \mathcal{D}_{\mathcal{X}}^t} \left\{ f(s^{({\bx})}_*, {\bx}) -  f(\hat{s}_t^{({\bx})}, {\bx}) \right\}.
\end{equation}
We again proceed to split the analysis into two cases.
\subsubsection{Regret for $(s^{(\underline{\bx}_t)}_*, \underline{\bx}_t) \notin S_t$}
First, we consider $(s^{(\underline{\bx}_t)}_*, \underline{\bx}_t) \notin S_t$. In this case, 
\begin{align}
    f(s^{(\underline{\bx}_t)}_*, \underline{\bx}_t) -  f(s_t^{(\underline{\bx}_t)}, \underline{\bx}_t) & \leq L_f|s^{(\underline{\bx}_t)}_* - s_t^{(\underline{\bx}_t)}| \qquad \text{(by definition of $L_f$)} \\
    & \leq 2(L_f/L_g')\beta_t^g\sigma^g_{t-1}(s_t^{(\underline{\bx}_t)}, \underline{\bx}_t) \qquad \text{(by \eqref{eqn:s*_stx*_bound_ii})}. \label{eq:add_iii}
\end{align}

Next, we derive the following:
\begin{align}
    f(s_t^{(\underline{\bx}_t)}, \underline{\bx}_t) - f(\hat{s}_t^{(\underline{\bx}_t)}, \underline{\bx}_t)
    & \leq \UCB^f_{t-1}(s_t^{(\underline{\bx}_t)}, \underline{\bx}_t) - \LCB^f_{t-1}(\hat{s}_t^{(\underline{\bx}_t)}, \underline{\bx}_t) \\ 
    & \leq \UCB^f_{t-1}(s_t^{(\underline{\bx}_t)}, \underline{\bx}_t) - \UCB^f_{t-1}(\hat{s}_t^{(\underline{\bx}_t)}, \underline{\bx}_t) + 2\beta^f_t\sigma^f_{t-1}(\hat{s}_t^{(\underline{\bx}_t)}, \underline{\bx}_t) \\ 
    & \leq 2\beta^f_t\sigma^f_{t-1}(\hat{s}_t^{(\underline{\bx}_t)}, \underline{\bx}_t) \qquad \text{(by \eqref{eqn:hat_s_t^x})}. \label{eq:add_iv}
\end{align}

Adding \eqref{eq:add_iii} and \eqref{eq:add_iv}, we have a bound on the worst-case simple regret incurred at round $t$ as follows:
\begin{align}
    f(s^{(\underline{\bx}_t)}_*, \underline{\bx}_t) - f(\hat{s}_t^{(\underline{\bx}_t)}, \underline{\bx}_t) &\leq 2(L_f/L_g')\beta_t^g\sigma^g_{t-1}(s_t^{(\underline{\bx}_t)}, \underline{\bx}_t) + 2\beta^f_t\sigma^f_{t-1}(\hat{s}_t^{(\underline{\bx}_t)}, \underline{\bx}_t).
\end{align}

\subsubsection{Regret for $(s^{(\underline{\bx}_t)}_*, \underline{\bx}_t) \in S_t$}
When $(s^{(\underline{\bx}_t)}_*, \underline{\bx}_t) \in S_t$, we can bound the instantaneous regret $r^{\mathcal{X}}_t$ directly, as follows:
\begin{align}
    r^{\mathcal{X}}_t = f(s^{(\underline{\bx}_t)}_*, \underline{\bx}_t) - f(\hat{s}_t^{(\underline{\bx}_t)}, \underline{\bx}_t) & \leq \UCB^f_{t-1}(s^{(\underline{\bx}_t)}_*, \underline{\bx}_t) - \LCB^f_{t-1}(\hat{s}_t^{(\underline{\bx}_t)}, \underline{\bx}_t) \\ 
    & \leq \UCB^f_{t-1}(s^{(\underline{\bx}_t)}_*, \underline{\bx}_t) - \UCB^f_{t-1}(\hat{s}_t^{(\underline{\bx}_t)}, \underline{\bx}_t) + 2\beta^f_t\sigma^f_{t-1}(\hat{s}_t^{(\underline{\bx}_t)}, \underline{\bx}_t) \\ 
    & \leq 2\beta^f_t\sigma^f_{t-1}(\hat{s}_t^{(\underline{\bx}_t)}, \underline{\bx}_t)  \qquad \text{(by \eqref{eqn:hat_s_t^x})}.
\end{align}

Therefore, considering the cumulative worst-case simple regret for $t = 1, \dots, T$, we have:
\begin{align}
    R_T^{\mathcal{X}} =\sum_{t=1}^T r^{\mathcal{X}}_t & = \sum_{t=1}^T \left(f(s^{(\bx_t)}_*, \bx_t) -  f(\hat{s}_t, \bx_t) \right) \\
    & \leq 2(L_f/L_g')\beta_t^g\sigma^g_{t-1}(s_t^{(\underline{\bx}_t)}, \underline{\bx}_t) + 2\beta^f_t\sigma^f_{t-1}(\hat{s}_t^{(\underline{\bx}_t)}, \underline{\bx}_t) \\ 
    & \leq 2\left((L_f/L_g') + 1\right) \max\{\beta_t^g \sigma_{t-1}^g(s_t, \bx_t), \beta_t^f \sigma_{t-1}^f(s_t, \bx_t)\}.
\end{align}

Once again, we can use Lemma 4 from \citep{chowdhury2017kernelized} to conclude that with probability at least $1-\delta$,
\begin{equation}
R_T^{\mathcal{X}} = O\left(\left(1+\frac{L_f}{L'_g}\right) \left(\beta^g_T\sqrt{T\gamma^g_T} + \beta^f_T\sqrt{T\gamma^f_T} \right) \right).
\end{equation}

\subsection{Refining the Regret Bounds} \label{sec:refined_regret}
In this section, we discuss an alternative acquisition function, defined as:
\begin{equation}
    \mathtt{acq}_t(s,\bx) = \max\{\beta^f_t \sigma^f_{t-1}(s,\bx), (L_f/L_g') \beta^g_t \sigma^g_{t-1}(s,\bx)\}, \text{, if } (s,\bx) \in G_t.
\end{equation}
Note that we only redefine the function for $(s,\bx) \in G_t$, while the function remains unchanged for the other cases (as stated in \eqref{eqn:acq_i} and \eqref{eqn:acq_iii}). This acquisition function provides certain desirable scaling properties of our theoretical guarantees, as discussed below. 

We note that the proofs of all theorems eventually used the acquisition function for establishing bounds of the following form (such as in \eqref{eqn:f_bound_2}, \eqref{eqn:inst_regret_ii} and \eqref{eqn:case_3_regret_ii}):
\begin{align}
    \beta^f_t \sigma^f_{t-1}(\mathbf{z}) &\leq \max\{\beta^f_t \sigma^f_{t-1}(s_t,\bx_t), \beta^f_t \sigma^f_{t-1}(s_t,\bx_t)\}, \\ 
    (L_f/L_g')\beta^g_t \sigma^g_{t-1}(\mathbf{z}) &\leq (L_f/L_g')\max\{\beta^f_t \sigma^f_{t-1}(s_t,\bx_t), \beta^f_t \sigma^f_{t-1}(s_t,\bx_t)\},
\end{align}
where $\mathbf{z}$ denotes some safe action (depending on the corresponding equation in our earlier proofs), while the other steps of the proofs did not depend on the acquisition function's definition. The $\max\{\beta^f_t \sigma^f_{t-1}(s_t,\bx_t), \beta^f_t \sigma^f_{t-1}(s_t,\bx_t)\}$ terms were then upper bounded by their sum (such as in \eqref{eqn:mod_acq_ref_i}) as follows:
\begin{align}
    \max\{\beta^f_t \sigma^f_{t-1}(s_t,\bx_t), \beta^g_t \sigma^g_{t-1}(s_t,\bx_t)\} &\leq \beta^f_t \sigma^f_{t-1}(s_t,\bx_t) + \beta^g_t \sigma^g_{t-1}(s_t,\bx_t), \\ 
    (L_f/L_g')\max\{\beta^f_t \sigma^f_{t-1}(s_t,\bx_t), \beta^g_t \sigma^g_{t-1}(s_t,\bx_t)\} &\leq (L_f/L_g')\beta^f_t \sigma^f_{t-1}(s_t,\bx_t) + (L_f/L_g')\beta^g_t \sigma^g_{t-1}(s_t,\bx_t).
\end{align}
Note that these steps introduced the factor $(L_f/L_g')$ into the $f$-terms as well, resulting in the regret bounds stated in our theorems.

With the alternative acquisition function given in \eqref{eqn:modified_acq}, our proofs can instead use the following refined steps:
\begin{align}
    \beta^f_t \sigma^f_{t-1}(\mathbf{z}) \leq \max\{\beta^f_t \sigma^f_{t-1}(s_t,\bx_t), (L_f/L_g')\beta^g_t \sigma^g_{t-1}(s_t,\bx_t)\}, \\
    (L_f/L_g')\beta^g_t \sigma^g_{t-1}(\mathbf{z}) \leq \max\{\beta^f_t \sigma^f_{t-1}(s_t,\bx_t), (L_f/L_g')\beta^g_t \sigma^g_{t-1}(s_t,\bx_t)\}.
\end{align}
Again using the fact that the maximum of the two terms is upper bounded by their sum, we can derive the following:
\begin{align}
    \beta^f_t \sigma^f_{t-1}(\mathbf{z}) \leq \beta^f_t \sigma^f_{t-1}(s_t,\bx_t) + (L_f/L_g')\beta^g_t \sigma^g_{t-1}(s_t,\bx_t), \\
    (L_f/L_g')\beta^g_t \sigma^g_{t-1}(\mathbf{z}) \leq \beta^f_t \sigma^f_{t-1}(s_t,\bx_t) + (L_f/L_g')\beta^g_t \sigma^g_{t-1}(s_t,\bx_t).
\end{align}

Therefore, when applying Lemma 4 from \citep{chowdhury2017kernelized}, the $(L_f/L_g')$ factor only gets multiplied with the $\beta_T^g \sqrt{T \gamma_T^g}$ term, and not the $\beta_T^f \sqrt{T \gamma_T^f}$ term. Hence, we get the following regret bound for $R_T$ in case 1 and case 3 (similarly for $R_T'$ and $R_T^{\mathcal{X}}$ in case 2):
\begin{equation}
    R_T = O\left(\beta_T^f \sqrt{T \gamma_T^f} + \frac{L_f}{L_g'}\beta_T^g \sqrt{T \gamma_T^g} \right).
\end{equation}

\paragraph{Behavior with respect to rescaling:} To motivate the notion of rescaling $f$ and/or $g$, we argue a certain equivalence between the following two problems for arbitrary $c > 0$:
\begin{itemize}
    \item[(i)] function $f$ with RKHS norm $B_f$ and noise level $R_f$;
    \item[(ii)] function $c f$ with RKHS norm $c B_f$ and noise level $c R_f$.
\end{itemize}
The equivalence comes from the fact that observations $y_t^f$ for problem (ii) can be obtained from those for problem (i) by simply multiplying by $c$ (or vice versa by dividing).  Hence, any algorithm for problem (ii) can be applied to problem (i), and vice versa.  The only difference is that in problem (ii) the regret is scaled by $c$ compared to problem (i).  In contrast, if $g$, $B_g$, and $R_g$ are similarly scaled (as well as $h$) then a similar equivalence holds without any change in the regret (which is measured only with respect to $f$).

In accordance with this discussion, a regret bound should ideally scale linearly when $f$ (and $B_f$, $R_f$) is scaled, and should stay unchanged when $g$ is scaled. This behavior is indeed observed in the refined regret bounds that we derived above: Scaling $f$ by $c$ results in scaling $\beta^f_T$ and $L_f$ by $c$, thereby scaling the regret bound by $c$ as well. However, scaling $g$ scales $\beta^g_T$ and $L_g'$ by $c$, and thus, the regret bound remains unchanged due to cancellation of the factor $c$. Although this scaling property of the regret bounds holds with the modified acquisition function, it comes at the cost of an additional dependence of the algorithm on $L_f$ and $L_g'$. Since this may not be desirable in practice due to unavailability of good estimates of these constants, we use the original forms of the acquisition in our theorems and experiments.

\subsection{Recovering the Guarantees of M-SAFEUCB} \label{sec:app_msafeucb}

As noted in Section \ref{sec:algo} under Case 3, modifying the functions $\mathtt{elim}_t$, $\mathtt{expd}_t$ and $\mathtt{acq}_t$ suitably gives us the \msafeucb algorithm of \citet{losalka2023benefits} for the goal of finding the optimal $s$ for every $\bx \in \dDX$, when both $f$ and $g$ are known to be monotone with respect to $s$.  Specifically, Theorem 1 in \citep{losalka2023benefits} gives a bound on a suitably-defined notion of cumulative regret, and Theorem 2 in \citep{losalka2023benefits} states that the entire safe boundary will be accurately identified after a certain number of rounds.  

To achieve this goal, we set $\mathtt{elim}_t(\bx) = \mathtt{false}$ for every $\bx$, set $\mathtt{expd}_t(\bx) = \mathtt{true}$ for every $\bx$ (except when $s^{(\bx)}_t = 1$), and set the acquisition function to be the same as that in \eqref{eqn:acq_iii}, except that here we only use $\sigma^g_{t-1}$ here and omit $\sigma^f_{t-1}$. Note that these choices imply that the algorithm does not need to use the constants $L_f$ and $L_g'$. 

We suppose that the algorithm is run using only the safety function $g$, which is reasonable since for each $\bx$, the best $s$ for $f$ is the same as the best $s$ for $g$ (since both are monotone), namely, $s_*^{(\bx)} = \max\{ s \in \dDS \,:\,g(s,\bx) \le h\}$.  Furthermore, since we do not consider $f$ in the algorithm in this case, we define $\hat{s}_t^{(\bx)}$ as the safe maximizer of ${\UCB}^g_{t-1}(\cdot, \bx)$ instead of $\UCB^f_{t-1}(\cdot, \bx)$.

\paragraph{Attaining Theorem 1 of \citep{losalka2023benefits}.}
Referring back to the proof for Theorem \ref{theorem:regret_bound_R_T_ii}, \eqref{eqn:g_safe_ii} holds for all $\bx \in \dDX$ by the definition of the optimal safe $s$, i.e., $s_*^{(\bx)}$. Also, note that $(s^{(\bx)}_*, \bx) \notin S_t$ in this case for any $\bx \in \dDX$ (when the confidence bounds are valid), since the optimal actions lie on the ``safe boundary'' (due to the monotonicity of both $f$ and $g$). Therefore, the following can be derived by similar reasoning as that used in \eqref{eqn:s*_stx*_bound_ii}:  
\begin{align}
    r^g_t &:= g(s^{(\bx_t)}_*, \bx_t) - g(s_t, \bx_t) \nonumber \\ 
    & \leq h - \LCB_{t-1}^g(s_t, \bx_t) \qquad \text{ (by \eqref{eqn:g_safe_ii})} \nonumber\\ 
    &\leq \UCB_{t-1}^g(s_t, \bx_t) - \LCB_{t-1}^g(s_t, \bx_t) \qquad \text{ (by \eqref{eqn:ucbg_xt_geq_h}} \nonumber\\ 
    &= 2\beta_t^g \sigma_{t-1}^g(s_t, \bx_t),
\end{align}
where $r^g_t$ is the instantaneous regret incurred by the algorithm by choosing $(s_t, \bx_t)$ at round $t$. Note that $s_t = s_t^{(\bx_t)}$ for every $t$ in this case (since all actions must be chosen from the set $G_t$), and ${\UCB}_{t-1}^g(s_t, \bx_t) < h$ does not hold (since any $(s_t^{(\bx)},\bx)$ such that $s_t^{(\bx)} = 1$ is never selected with the redefined $\mathtt{expd}_t$). Also note that the problem setting of \citep{losalka2023benefits} precludes the scenario where $g$ is safe over the entire input domain $\dD$; thus, excluding actions with $s_t^{(\bx)}=1$ does not lead to the situation that the algorithm is unable to chose any action in some round $t$.

In the case that $f = g$, we readily obtain 
Theorem 1 from \citep{losalka2023benefits} by summing $r^g_t$ over $t = 1, \dots, T$, and using Lemma 4 from \citep{chowdhury2017kernelized}.  Similar to the setup in \citep{losalka2023benefits}, this derivation does not require $L_g' > 0$ (i.e., $g$ only needs to be non-decreasing rather than strictly increasing in this case).   
A similar argument also applies in the case that $f \ne g$ and both $f$ and $g$ are monotone with respect to $s$, but in this more general case, we need to that assume $L_g' > 0$ (and $L_f < \infty$) for the reasons discussed in Section \ref{app:stuck}, and the final regret bound incurs a $1+\frac{L_f}{L'_g}$ factor in the same way as Theorems \ref{theorem:regret_bound}--\ref{theorem:regret_bound_D_T_ii}.  The algorithm itself does not need to know $L_f$ and $L’_g$.

\paragraph{Attaining Theorem 2 of \citep{losalka2023benefits}.}
We refer to the proof of our Theorem \ref{theorem:regret_bound_D_T_ii} and consider the regret analysis for $(s^{(\underline{\bx}_t)}_*, \underline{\bx}_t) \notin S_t$ only (for the same reason as above). Note that given the design of the algorithm, we have $s_t = s_t^{(\bx)} = \hat{s}_t^{(\bx)}$, i.e., for any $\bx$, the $s$ on the current ``safe boundary'' is also the one that maximizes ${\UCB}^g_{t-1}$.

Furthermore, we note that if $s^{(\bx)}_t =1$ for any $\bx \in \dDX$, that would imply that for this specific value of $\bx$, $(s,\bx)$ is safe for every $s \in \dDS$ (based on validity of our confidence bounds).  Accordingly, in the following, we only consider $\bx \in \dDX$ for which $s^{(\bx)}_t < 1$. 

Based on the discussion above, the worst-case (with respect to $\bx$) suboptimality of the algorithm's best guess (of the optimal $s$) can be bounded as follows:
\begin{align}
\label{eq:m-safeucb_ii}
     \max_{\bx: s^{(\bx)}_t < 1}& \left\{g(s^{(\bx)}_*, \bx) - g(\hat{s}^{(\bx)}_t, \bx)\right\} \nonumber \\ 
     & \leq \max_{\bx: s^{(\bx)}_t < 1} \left\{h - \LCB(\hat{s}^{(\bx)}_t, \bx)\right\} \nonumber \\ 
     & \leq \max_{\bx:s^{(\bx)}_t < 1} \left\{\underline{\UCB}({s}^{(\bx)}_t, \bx) - \LCB(\hat{s}^{(\bx)}_t, \bx) \right\} \quad \text{ (since $\underline{\UCB}({s}^{(\bx)}_t, \bx) \ge h$ when ${s}^{(\bx)}_t < 1$)} \nonumber \\
     & \leq \max_{\bx:s^{(\bx)}_t < 1}\left\{ 2\beta^g_t\sigma^g_{t-1}({s}^{(\bx)}_t, \bx)\right\} \quad \text{ (since $  \hat{s}_t^{(\bx)} = s_t^{(\bx)}$, and $\underline{\UCB}({s}^{(\bx)}_t, \bx) \le {\UCB}({s}^{(\bx)}_t, \bx)$}) \nonumber \\
     & = 2\beta^g_t\sigma^g_{t-1}({s}_t, \bx_t) \quad \text{ (since $\mathtt{acq}_t$ maximizes $\sigma^g_{t-1}(s,\bx)$ over $(s,\bx) \in G_t$)}.
\end{align}

As earlier, applying Lemma 4 from \citet{chowdhury2017kernelized} upper bounds the cumulative value of the worst-case regret derived in \eqref{eq:m-safeucb_ii}.  Doing so essentially recovers Theorem 2 of \citet{losalka2023benefits} (with $f = g$), with the difference that their result is not based on a cumulative measure, but rather based on returning the best estimate of $s$ for each $\bx$ \emph{after} all queries have been taken.  However, the latter follows as a simple consequence of the former by letting the final $\hat{s}^{(\bx)}_t$ be the maximum among $\{\hat{s}^{(\bx)}_t\}_{t=1}^T$ (and thus the one with the lowest regret), using the fact that the minimum regret is no higher than the average (with respect to $t \sim {\rm Uniform}(1,\dotsc,T)$), and noting that such an average is precisely $\frac{1}{T}$ times the cumulative regret.

\begin{figure*}[t!]
  \centering
  \setlength\tabcolsep{2pt}
  \begin{tabular}{cc}
     \includegraphics[width=0.5\linewidth]{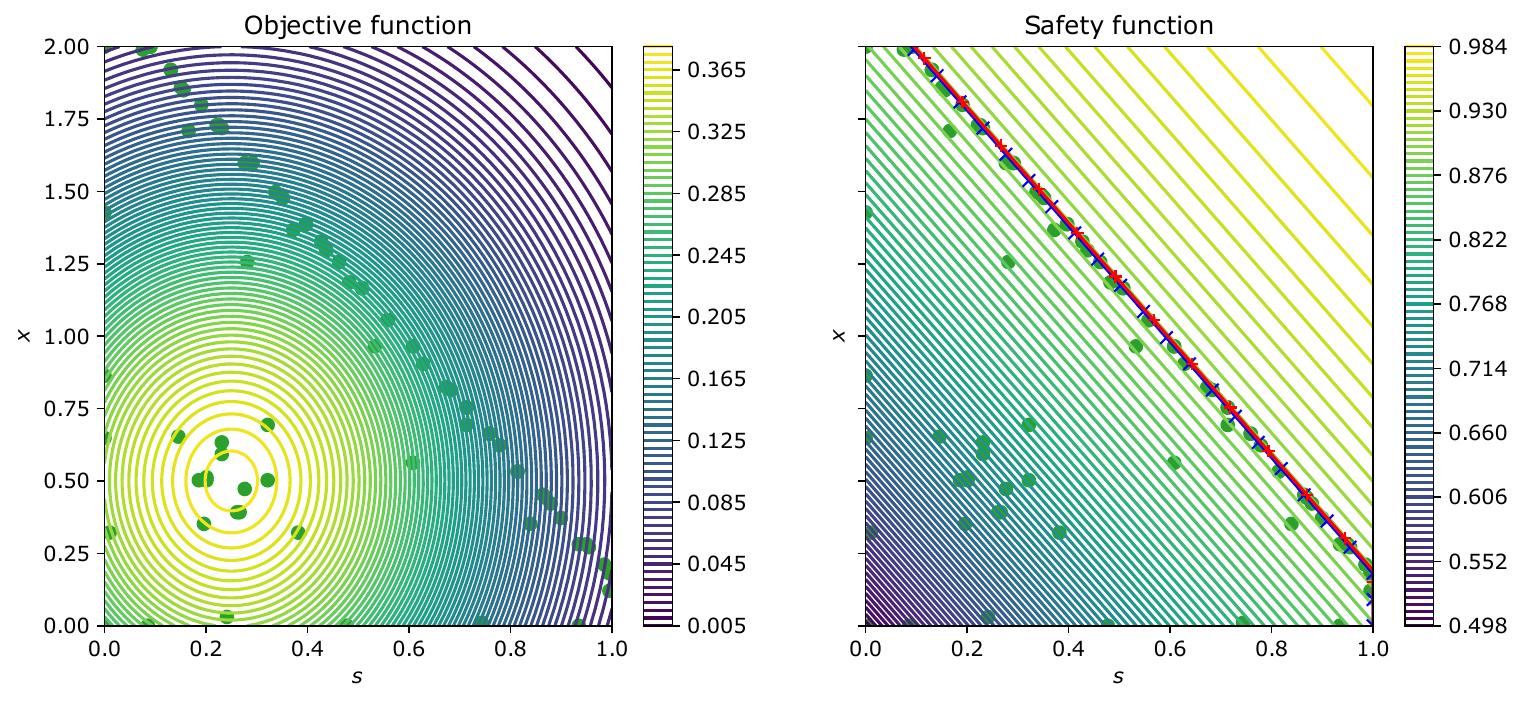} &
     \includegraphics[width=0.5\linewidth]{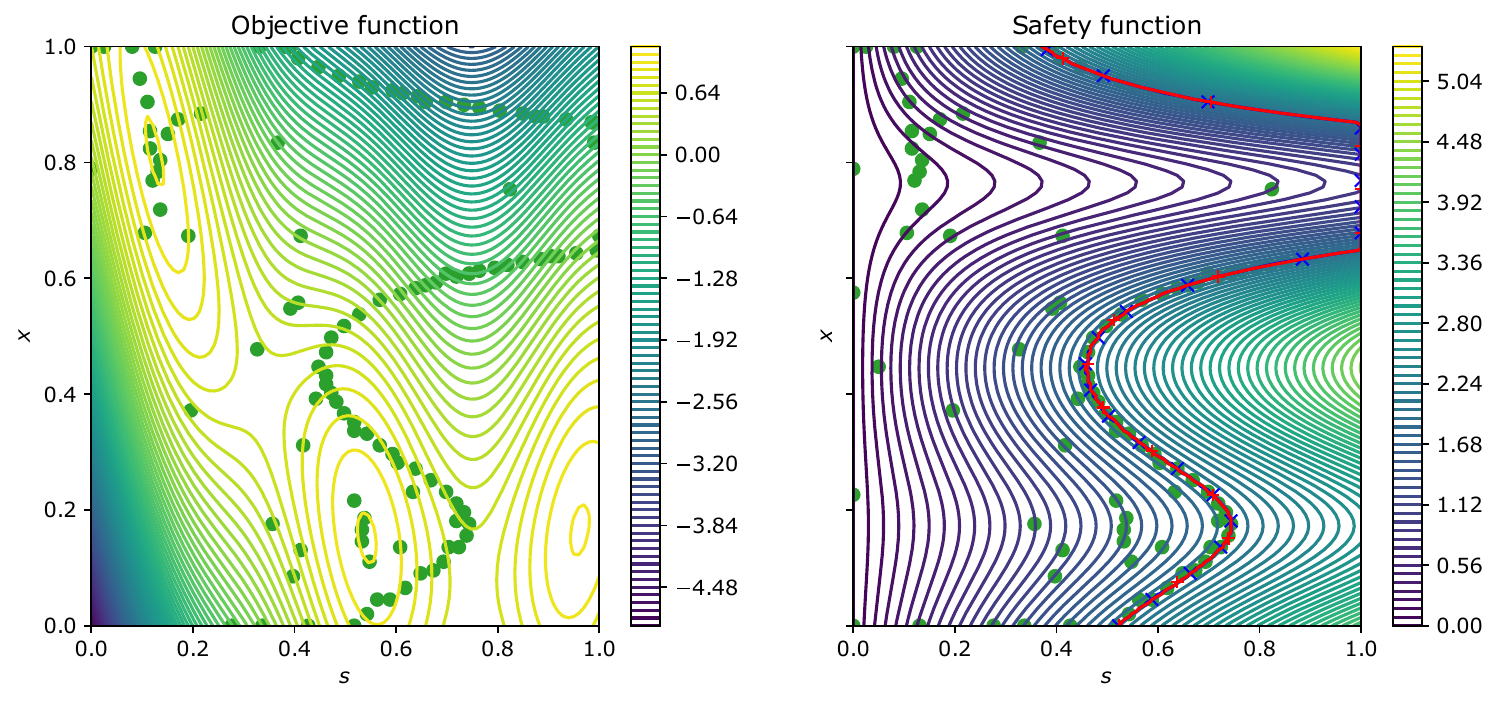}\\
      \includegraphics[width=0.5\linewidth]{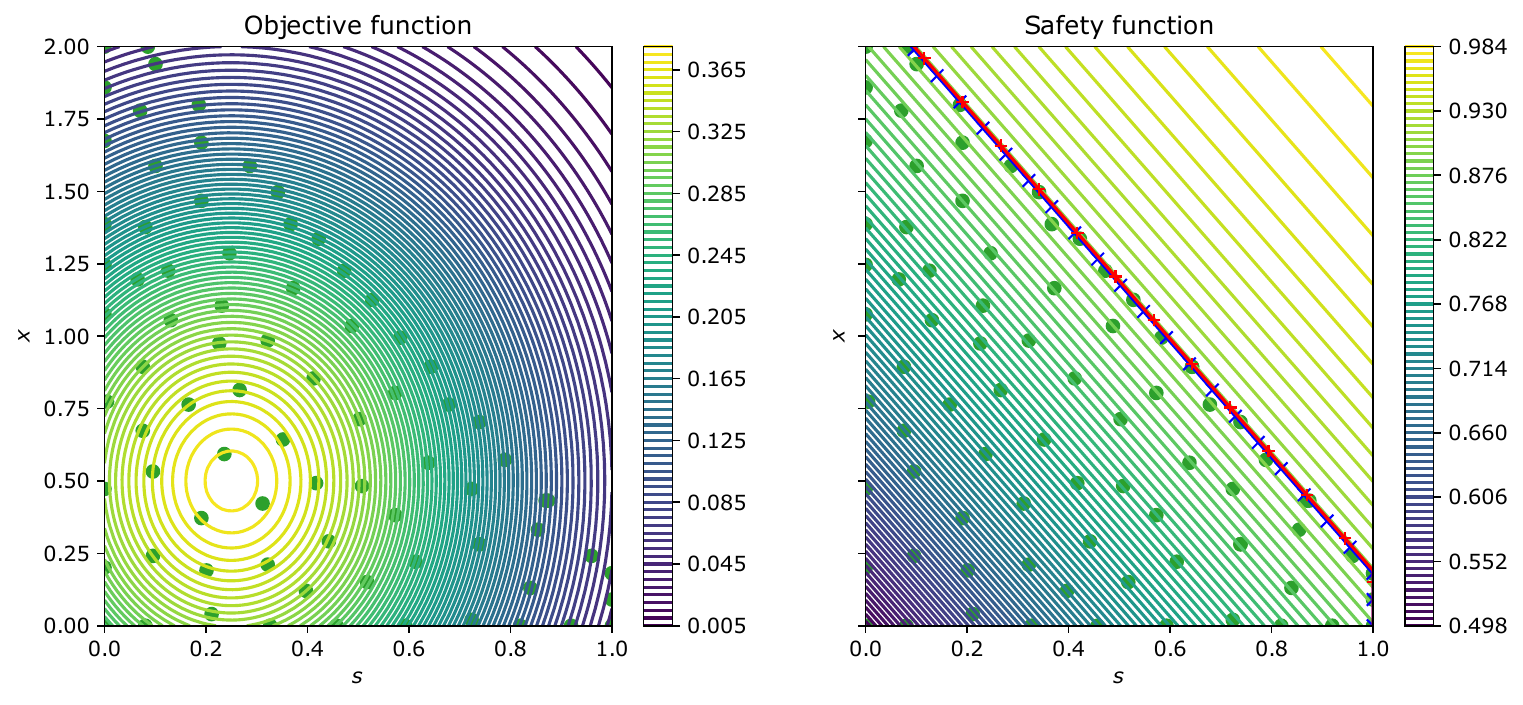} &
     \includegraphics[width=0.5\linewidth]{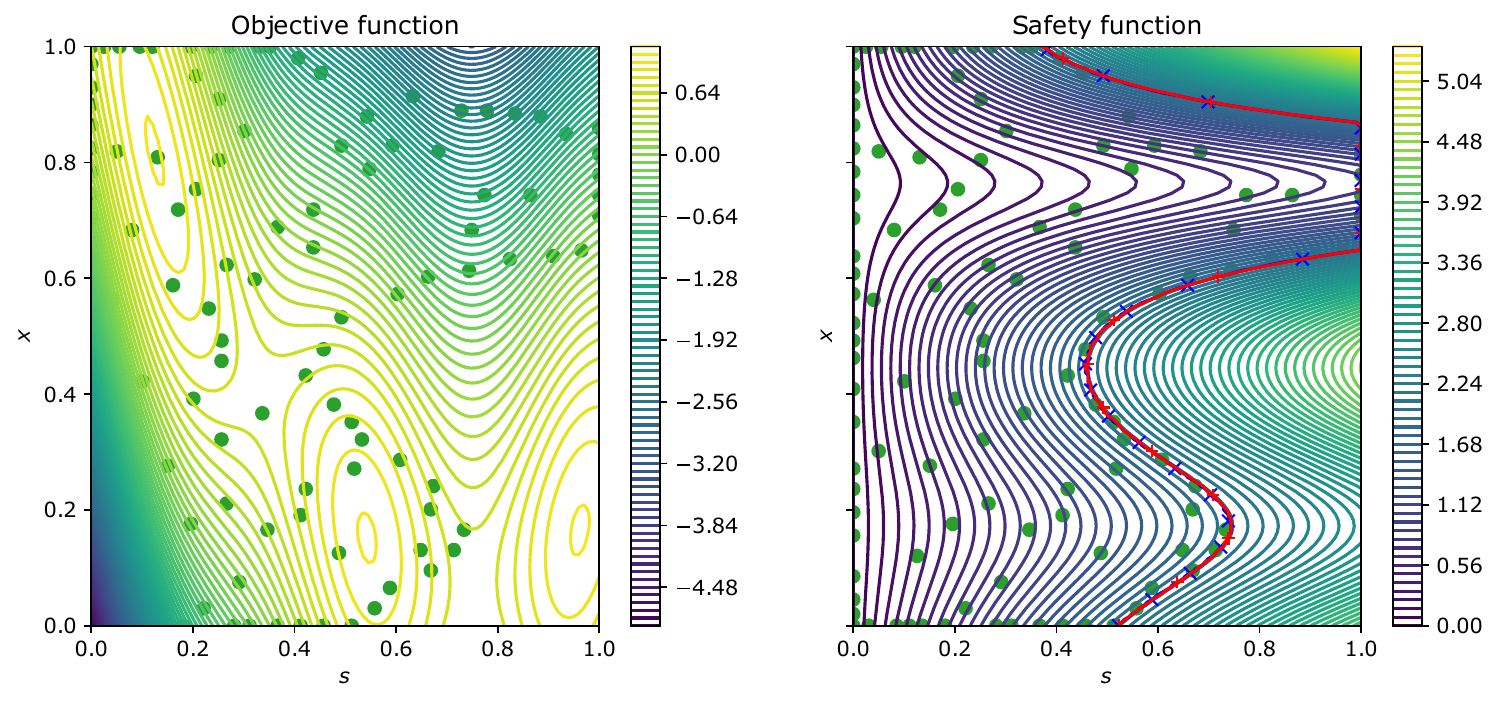}
     \end{tabular}
  \caption{The first two columns shows the actions sampled by the \safeopt (row 1) and \predvar (row 2) algorithms for the simulated clinical trial experiment, while the last two columns shows the corresponding plots for the synthetic 2D experiment (from Section \ref{sec:exp}). The true safe boundary is shown in red and the boundary discovered by the algorithm is shown in blue (in columns 2 and 4).}
\label{fig:bo_plots_app}
\end{figure*}

\section{ADDITIONAL EXPERIMENTS AND DETAILS}

We first provide additional experimental results in Section \ref{sec:app_more_results}.  The additional experimental details (e.g., choice of kernel and description of baselines) are deferred to Section \ref{sec:app_exp_details}.

\subsection{Additional Experimental Results} \label{sec:app_more_results}

\paragraph{Actions Sampled by Baselines.} In this section, we first present the actions sampled by \safeoptmc and \predvar in Figure \ref{fig:bo_plots_app} (similar to Figure \ref{fig:bo_plots}, which shows the actions selected by \msafeoptns). While \predvar samples actions throughout the safe set in order to reduce uncertainty with respect to both $f$ and $g$, \safeoptmc samples either close to the ``safe boundary'' (potential expanders) or among the potential maximizers. See Appendix \ref{app:results_discussion} for further discussion on the performance of these two algorithms, and how they compare with \msafeoptns.

\begin{figure*}[t!]
  \centering
  \setlength\tabcolsep{2pt}
  \begin{tabular}{ccc}
     \includegraphics[width=0.32\linewidth]{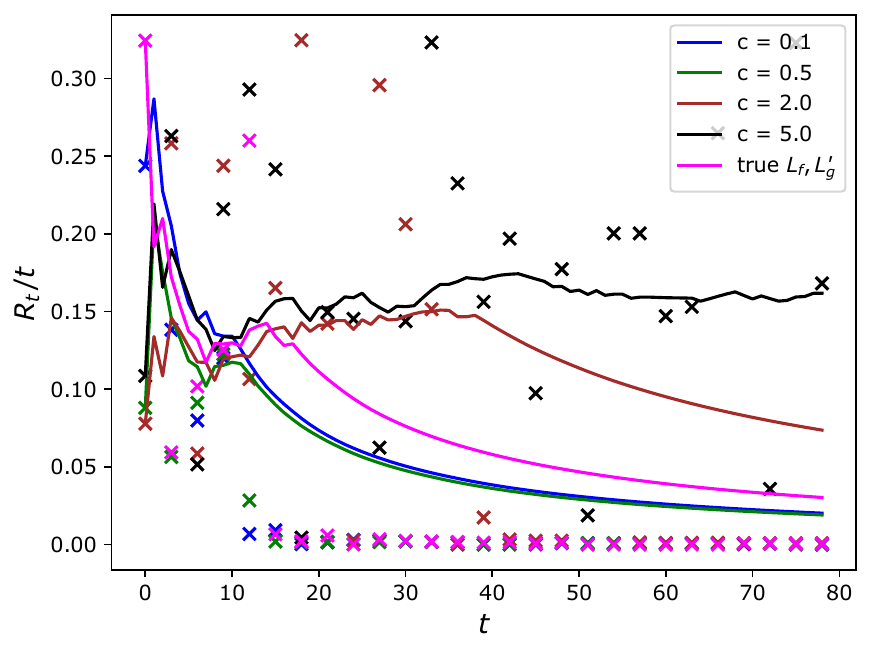} &
     \includegraphics[width=0.32\linewidth]{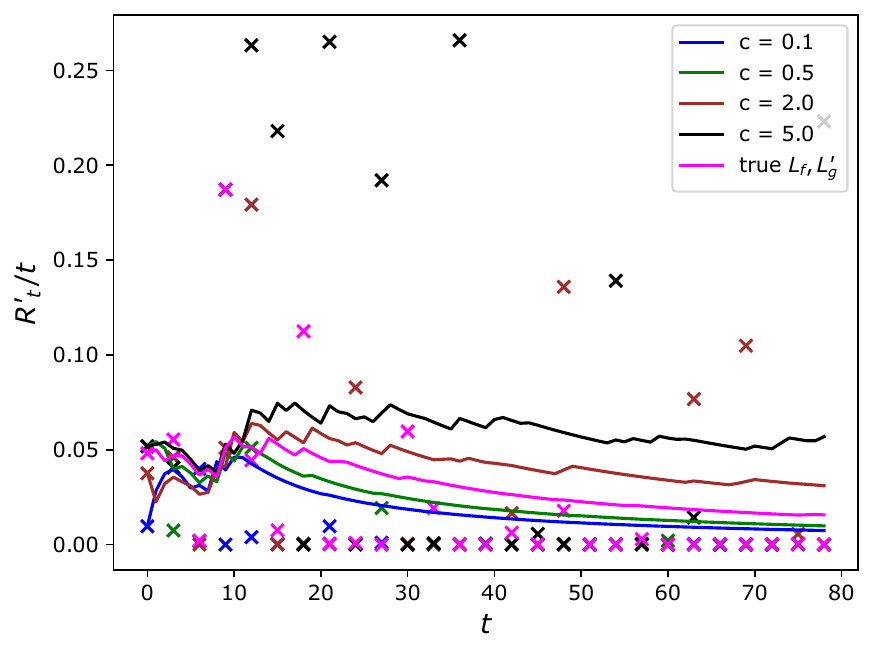} &
     \includegraphics[width=0.32\linewidth]{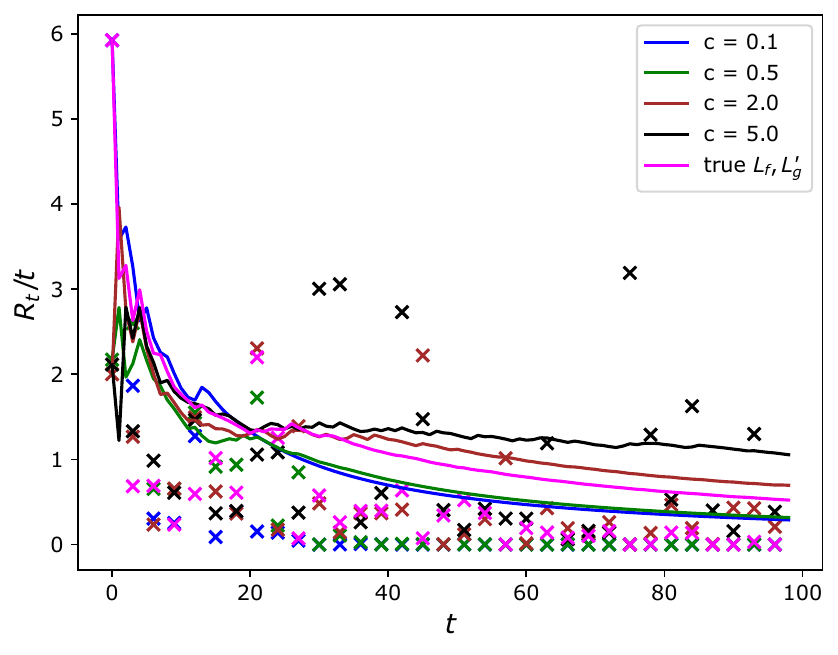} 
     \end{tabular}
  \caption{The first column shows the normalized cumulative regret, $R_t/t$, incurred by \msafeopt for the simulated clinical trial experiment for different values of $c$ (where $L_f$ is set to $cL_f$, and $L_g'$ is set to $L_g'/c$), while the second column shows that for $R'_t/t$. The third column corresponds to the synthetic 2D experiment (showing $R_t/t$). The instantaneous regret values are shown using markers.\\ 
  }
\label{fig:lf_lg}
\end{figure*}

\paragraph{Effect of $L_f$ and $L'_g$: } In this section, we evaluate the performance of \msafeopt in the scenario that the parameters $L_f$ and $L_g'$ used in the algorithm differ from the corresponding true values. As mentioned in Section \ref{sec:problem}, an overestimate of $L_f$ and an underestimate of $L_g'$ also allow our theoretical guarantees to hold (since \eqref{eqn:lf} and \eqref{eqn:lg'} remain valid). Therefore, we design this set of experiments by varying the values of $L_f$ and $L_g'$ as follows: $L_f \leftarrow cL_f, L_g' \leftarrow L_g'/c$ with $c \in \{2,5\}$. Additionally, to see how \msafeopt performs in the reverse scenario, i.e., $L_f$ is underestimated and $L_g'$ is overestimated, we also consider $c \in \{0.1, 0.5\}$.

Intuitively, $c>1$ implies that the algorithm becomes more cautious when setting $\mathtt{elim}_t = \mathtt{true}$ and $\mathtt{expd}_t = \mathtt{false}$, and thus, the regret is expected to converge slower than that with $c=1$. This is also observed in our experiments, as shown in Figure \ref{fig:lf_lg}.

On the other hand, $c<1$ leads to more aggressive elimination of $\bx$'s and expansion of actions on the ``safe boundary''. This may cause the regret to converge faster (as is also the case in Figure \ref{fig:lf_lg}). However, this may come at the expense of failing to explore potentially optimal regions due to improper elimination/non-expansion based on the incorrect estimates. In the present scenario with ``well-behaved'' objective and safety functions, this behavior tends to be avoided. 

We also note that it is possible to be more robust to such choices of $L_f$ and $L_g'$ ($c<1$), and also possibly imprecise knowledge of the kernel, by making the following change to the algorithm: eliminate $\bx$'s from the whole set $\dDX$ (instead of the previous set $\mathcal{D}^t_{\mathcal{X}}$) in each round, such that any $\bx$ that may have been incorrectly eliminated in a specific round still has a chance to be recovered once the confidence bounds are refined in subsequent rounds.  In view of this improved robustness, we adopt this strategy in our implementation. 

In the rest of this subsection, we compare the performance of \msafeopt with the baseline algorithms on two other problems -- one with a three dimensional input domain, and another being a modification of the pendulum swing-up problem, a classic control task \citep{brockman2016openai}. 

\begin{figure*}[t!]
  \centering
  \setlength\tabcolsep{2pt}
  \begin{tabular}{cc}
     \includegraphics[width=0.35\linewidth]{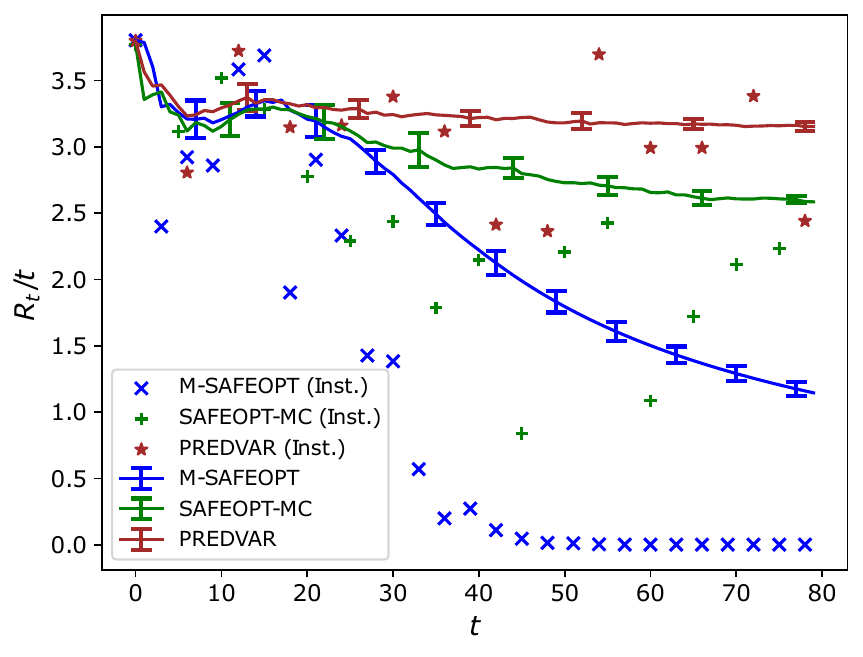} &
     \includegraphics[width=0.35\linewidth]{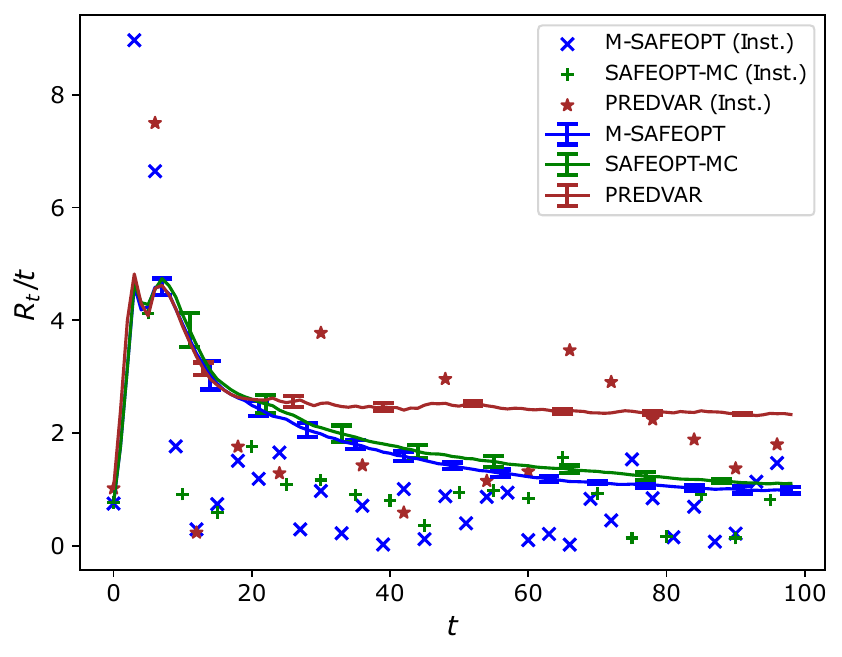}
     \end{tabular}
  \caption{The first column shows the normalized cumulative regret, $R_t/t$, for the synthetic 3D experiment, while the second column shows that for the pendulum swing-up experiment. The corresponding instantaneous regret values are shown using markers.\\ 
  }
\label{fig:regret_plots_app}
\end{figure*}

\paragraph{Synthetic 3D Functions.} For this experiment, we used the Hartmann-3 function \citep{hedar2013global} as the objective function $f_{\mathrm{syn}_{3D}}$, along with the following safety function:
\begin{equation}
    g_{\mathrm{syn}_{3D}}(s, \bx) = s + x_1^2+x_2^3,
\end{equation}
where $\bx = (x_1, x_2)$. The domain is set to be $[0,1]^3$, and is discretized into $75 \times 75 \times 75$ linearly spaced points. We set the safety threshold to $h=2$, such that $g_{\mathrm{syn}_{3D}}$ satisfies the assumptions of our problem setup, i.e., monotonicity with respect to $s$, and safety for $s=0$ for all values of $\bx \in \dDX$. The function evaluations are set to be noiseless. 

The results of running \msafeoptns, along with the baseline algorithms, are presented in Figure \ref{fig:regret_plots_app}.  Similar to our previous experiments, we see that \msafeopt is able to converge to the global safe optimum while attaining sublinear regret, whereas the regret incurred by \safeoptmc and \predvar is not sublinear, since they continue to explore suboptimal regions.

\paragraph{Pendulum Swing-up Problem.}
For this problem, we consider the pendulum swing-up problem, which is a classic control problem available as an OpenAI Gym environment \citep{brockman2016openai}. The task is to apply torque to the free end of a pendulum such that it swings and stays in the upright position, i.e., it attains an angular velocity of zero when it reaches this position. We adapt the problem to our setup as follows. The initial angle ($\bx$) of the pendulum lies within the range $\dDX = [-2\pi + \pi/36, -\pi -\pi/36]$ (with angle $= 0$ denoting the upright position), and the torque ($s$) lies in $\dDS = [0,1]$ ($s$ is scaled up by a factor of $40$ to calculate the motion, so that the upright position is attainable). The torque is applied only once at the beginning, while the episode is run for $100$ time steps. 

The reward function is defined as follows (similar to \citep{losalka2023benefits}):
\begin{gather}
    f_n(s,\bx) = 
    \begin{cases}
    -\theta_n^2(s,\bx) -\frac{\Dot{\theta}_n^2(s,\bx)}{10} - \frac{s^2}{1000}, & \text{ if } \theta_n(s,\bx) \leq 0 \\
    -\Dot{\theta}_{up}(s,\bx)  &\text{ if } \theta_n(s,\bx) > 0,
    \end{cases}
    \\[2mm]
    f(s,\bx) = \max_{n \leq 100} f_n(s,\bx),
\end{gather}
where $\theta_n(s,\bx)$ and $\Dot{\theta}_n(s,\bx)$ denote the angle and angular velocity of the pendulum at the $n^{th}$ time step, and $\Dot{\theta}_{up}(s,\bx)$ denotes the angular velocity of the pendulum when it crosses the upright position, starting with an initial angle and torque of $\bx$ and $s$ respectively.

The safety function is set to be equal to the maximum magnitude of the angular velocity attained by the pendulum at any time step during the episode: 
\begin{equation}
    g(s,\bx) = \max_{n \leq 100} |\Dot{\theta}_n(s,\bx)|,
\end{equation}
and the safety threshold is set to $h=9$.  In this experiment, we consider noisy queries for both $f$ and $g$, namely, additive $\mathcal{N}(0,0.05)$ noise.

The initial angular velocity is always set to zero such that our safety assumption is satisfied ($s=0$ being safe for every $\bx$); this is because the threshold magnitude of velocity is unattainable for any value of the initial angle ($\bx \in \dDX$), unless a positive torque ($s>0$) is applied. One episode (with $100$ time steps) is simulated to calculate the reward and safety values in every iteration of optimization. 

The input domain is discretized into $100 \times 100$ linearly spaced points, and the results of running \msafeoptns, along with the baseline algorithms are presented in Figure \ref{fig:regret_plots_app}.  In this case, \msafeopt and \safeoptmc behave similarly, whereas \predvar incurs higher regret due to its exploratory nature.

\begin{figure*}[t!]
  \centering
  \setlength\tabcolsep{2pt}
  \begin{tabular}{cc}
     \includegraphics[width=0.35\linewidth]{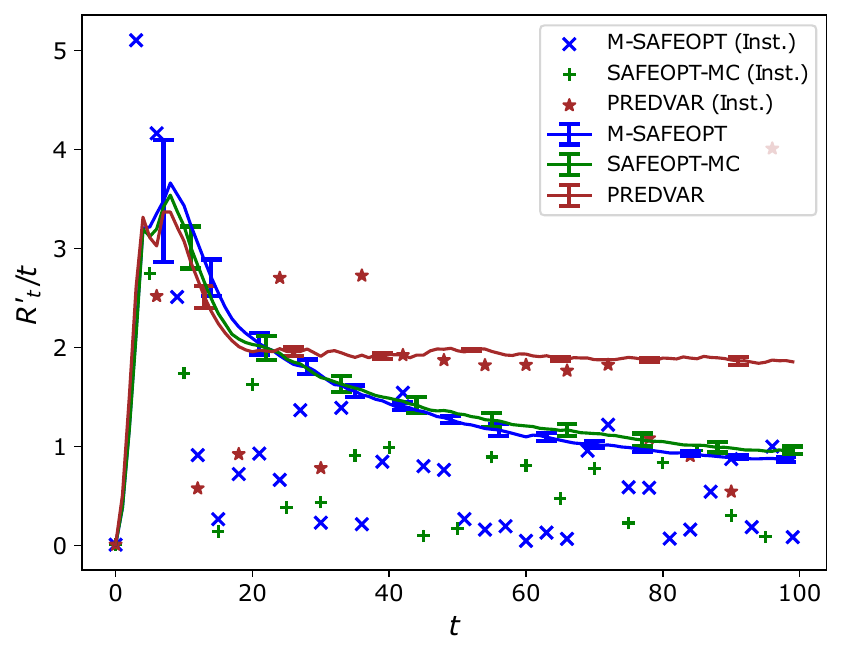} &
     \includegraphics[width=0.35\linewidth]{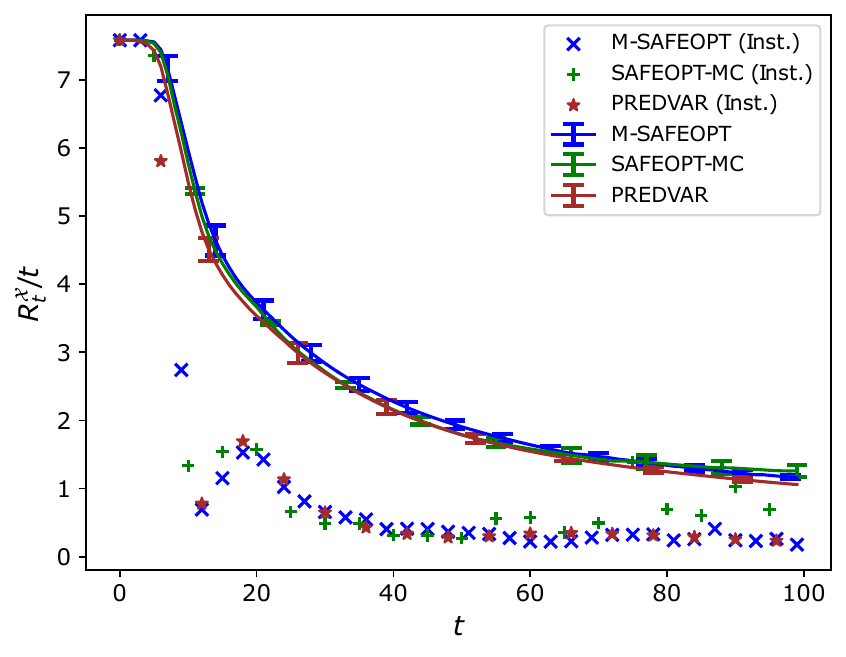}
     \end{tabular}
  \caption{The first column shows the normalized cumulative regret, $R_t'/t$, for the pendulum swing-up experiment, while the second column shows that for $R^{\mathcal{X}}_t/t$ when running \msafeopt (Case 2) along with the baseline algorithms. The corresponding instantaneous regret values are shown using markers.\\ 
  }
\label{fig:regret_plots_app_ii}
\end{figure*}
\subsubsection{Further Discussion} \label{app:results_discussion}
With respect to the performance of the baseline algorithms \safeoptmc and \predvar, we highlight the following observations:
\begin{itemize}
    \item \safeoptmc tries to simultaneously explore the potentially optimal regions ($M_t$) and actions that could potentially expand the safe set ($G_t$) as seen via the actions sampled in Figure \ref{fig:bo_plots_app}. However, it expands while being ``blind'' to the objective function $f$ (i.e., it does not try to evaluate the suboptimality of the potential expanders). This appears to be unavoidable in the general scenario, since without any known structure on $g$ (e.g., monotonicity), it is not possible to predict which region of the input domain may contain the optimal safe action; hence, the entire reachable safe set must be explored by \safeoptns.  This is reflected in the performance of \safeopt with respect to $R_t$ and $R'_t$ (e.g., see the first and second columns of Figure \ref{fig:regret_plots}, and the corresponding plots in Figure \ref{fig:regret_plots_app} and \ref{fig:regret_plots_app_ii}).  
    \item Due to its exploratory nature, \predvar performs well in terms of $R^{\mathcal{X}}_t$ (as observed in the third column in Figure \ref{fig:regret_plots}), i.e., it is able to identify a near-optimal $s$ for each $\bx \in \dDX$. However, for the same reason, it performs poorly with respect to $R_t$ and $R'_t$, i.e., it does not make progressively ``better'' choices (since it tries to reduce uncertainty over $f$ and $g$ over $S_t$ without considering potential optimality separately), as observed in the first and second columns of Figure \ref{fig:regret_plots}, as well as both plots in Figure \ref{fig:regret_plots_app} and the first column of Figure \ref{fig:regret_plots_app_ii}. 
\end{itemize}

\subsection{Details of Experiments} \label{sec:app_exp_details}
In this section, we describe the details of the setup of our experiments, including the details of the objective and safety functions, and the implementation details of the algorithms.

\subsubsection{Experimental Setup}

\paragraph{Simulated Clinical Trial.}
For this experiment, we consider the dose efficacy ($f_{\mathrm{eff}}$) and dose toxicity ($g_{\mathrm{tox}}$) functions as described in \eqref{eqn:dose_eff} and \eqref{eqn:dose_tox} respectively. Specifically, the values of $\theta_i$'s are set as follows: $\theta^f_0 = 1, \theta^f_1 = 2, \theta^f_2 = 1, \theta^f_3 = -4, \theta^f_4 = -1, \theta^g_1 = 2, \theta^g_2 = 1$ to give us the functions shown in Figure \ref{fig:bo_plots}. As stated in Section \ref{sec:exp}, we assume that $d_1$ and $d_2$ represent the dosages of two different drugs that are administered as a combination in a clinical trial. For drug combinations, it is common to observe non-monotonic behavior of the efficacy (e.g., immunotherapy trials \citep{cai2014bayesian}). Since the toxicity increases monotonically with respect to both $d_1$ and $d_2$ in this experiment, either could be treated as the safety variable $s$. We use $d_1$ as the safety variable in our implementation, and try to find the global safe optimal dose combination for goal (i) (as in Section \ref{sec:problem}). 

For goal (ii) (as in Section \ref{sec:problem}), the same functions are used in our experiment. Since we need to find the optimal safe $d_1$ for every $d_2$ in this case. From the perspective of practical relevance, it may be more sensible to interpret $d_2$ as the age of patients, so that the goal translates to finding the optimal dose ($d_1$) for every age ($d_2$). However, from  the perspective of evaluation of the algorithms, the proposed meaning/interpretation of each variable does not have any quantitative effect.

The input domain is set to be $[0,1] \times [0,2]$, which is discretized into $200\times 200$ linearly spaced points in the domain. The function evaluations observed by the algorithm are noiseless. The safety function  $g_{\mathrm{tox}}(s, \bx)$ satisfies the assumptions of our problem setup, i.e., strict monotonicity with respect to $s$, and safety at $s=0$ for all $\bx \in \dDX$. However, $f_{\mathrm{eff}}$ is non-monotonic in both $s$ and $\bx$.

\paragraph{Synthetic 2D Functions.}
We consider the following synthetic functions:
\begin{align}
    &f_{\mathrm{syn}_1} = \alpha\cdot\left((x - bs^2 + cs -6)^2 + 10(1-t)\cos(s) + \Delta\right),\\ 
    &\qquad g_{\mathrm{syn}_1} = 2s(e^y \sin(10y) + \sin(5y) + 5)/3, 
\end{align}
where the parameters are set to the following values: $\alpha = 1/51.95, \Delta = -44.81, b = 5.1/4\pi^2, c =5/\pi , t = 1/8\pi$ as per those used for defining the scaled Branin function \citep{picheny2013benchmark}. $y$ is set to $x + 1/3$  for defining $g_{\mathrm{syn}_1}$, so that one of the three local optima of the objective function gets excluded from the safe region. We use these functions due to the presence of multiple local optima (of $f_{\mathrm{syn}_1}$), less smooth optimization surfaces for both $f_{\mathrm{syn}_1}$ and $g_{\mathrm{syn}_1}$ (compared to $f_{\mathrm{eff}}$ and $g_{\mathrm{tox}}$), and difficulty in eliminating $\mathbf{x}$’s and/or terminating expansion due to presence of near-optimal actions in the unsafe region close to safe boundary. This makes it more difficult for the algorithm to identify and eliminate suboptimal regions of the input space. The function evaluations observed by the algorithm are noiseless. 

The input domain is set to be $[0,1] \times [0,1]$, and is discretized into $200\times 200$ linearly spaced points in the domain. The function evaluations observed by the algorithm are noiseless. The safety function  $g_{\mathrm{syn}_1}(s,\bx)$ satisfies strict monotonicity with respect to $s$, and safety at $s=0$ for all $\bx \in \dDX$.

\subsubsection{Implementation Details} \label{sec:implementation}

In all our experiments, $\beta_f^t$ and $\beta_g^t$ are set to a constant value of $3$ in all rounds $t \ge 1$. The use of a constant $\beta$ is fairly common in the Bayesian optimization/safe Bayesian optimization literature, since the theoretical values tend to be overly cautious to aid the derivation of theoretical guarantees. 

We use the Trieste library for Bayesian optimization in our implementations \citep{trieste2023}. All experiments are repeated $5$ times, and the plots in Figure \ref{fig:regret_plots} and \ref{fig:regret_plots_app} show the mean values of the corresponding notions of regret (both instantaneous and cumulative), along with the standard deviations via error bars. 

\paragraph{Gaussian Process Model.} In all our experiments, the Gaussian process models for both $f$ and $g$ use the Mat\'ern-$\frac{5}{2}$ kernel, with the length scales and variance of the kernel set to be trainable. A log-normal prior is used for both the variance and the length scales with a standard deviation $1$. The mean values for the length scales are set to $0.2$, and that for the variance is set to $1$ for the simulated clinical trial and the synthetic 3D experiments, and $9$ for the experiment with the synthetic 2D functions. The GP models assumes a low noise level of $10^{-5}$ for numerical stability (except for the pendulum swing-up experiment, where the algorithm observes noisy evaluations of $f$ and $g$, the noise being sampled from $\mathcal{N}(0,0.05)$ and the GP model assumes a noise level with variance $0.05$). Seeking minimal manual tuning, the choices of parameters mostly follow the recommendations in \citep{trieste2023}. 

\paragraph{M-SafeOpt Algorithm.} For implementing our \msafeopt algorithm, we first find the values of $L_f$ and $L_g'$ by computing the gradients of $f$ and $g$ over a finely discretized grid of points in the input space.  We directly use these values unless stated otherwise, but we recall that Appendix \ref{sec:app_more_results} also explores the algorithm's robustness to misspecified values.

We also note that while our theory holds for continuous $\bx$, our implementation relies on discretization of $\bx$ due to the explicit for-loop over $\bx$. For continuous domains, alternative approaches involving approximations are possible; however, we do not claim the validity of our theoretical guarantees for such variations. A few such alternatives are discussed in Appendix C.2 in \citep{losalka2023benefits} for the case that $f=g$, and similar approaches can also be adopted in our setting with distinct $f$ and $g$.

\paragraph{SafeOpt Algorithm.}  For implementing the \safeopt algorithm \citep{sui2015safe}, we specifically use the variant proposed by \citet{berkenkamp2021bayesian}, \safeoptmcns, that works with distinct objective and safety functions. To incorporate the knowledge of monotonicity of $g$ in the implementation, we explicitly define the set $G_t$ of \textit{potential expanders} as the set of actions on the current ``safe boundary'' as discovered by the algorithm and as defined by \msafeoptns. The only difference is that in this case, we remove any action with $s_t^{(\bx)} = 1$ from $G_t$; this is because \safeopt defines $G_t$ as the set of actions that could potentially expand the current safe set of actions, whereas $s_t^{(\bx)} = 1$ implies that $g(s,\bx)$ has been discovered to be safe for every $s \in \dDS$ already. On the other hand, the set of \textit{potential maximizers}, $M_t$, is computed exactly as defined in \citep{sui2015safe}. We avoid using Lipschitz constants, instead relying on the modification proposed by \citep{berkenkamp2017safe}.  

\paragraph{PredVar Algorithm.} For the \predvar algorithm, we conceptually rely on \citep{schreiter2015safe}, while extending the algorithm to consider multiple functions simultaneously. We define the the currently known safe region in the same way as that in \msafeoptns. The acquisition function uses the maximum among the width of the confidence intervals given by the GP models for $f$ and $g$ for all actions in the safe set $S_t$. Thus, \predvar behaves as a purely exploratory algorithm that tries to minimize variance across the safe input domain for both $f$ and $g$ simultaneously, while also expanding the safe set.

\section{DISCUSSION}

\subsection{Necessity of $L_f$ and $L'_g$} \label{app:stuck}

Recall that we assume a minimum growth rate of $g$ with respect to $s$ (i.e., $L'_g > 0$) and a a maximum growth rate of $f$ with respect to $s$ (i.e., $L_f < \infty$).  Here we argue that these assumptions (or similar) are in fact essential for attaining meaningful regret bounds in our setting.

\begin{figure}
    \begin{centering}
        \includegraphics[width=0.5\columnwidth]{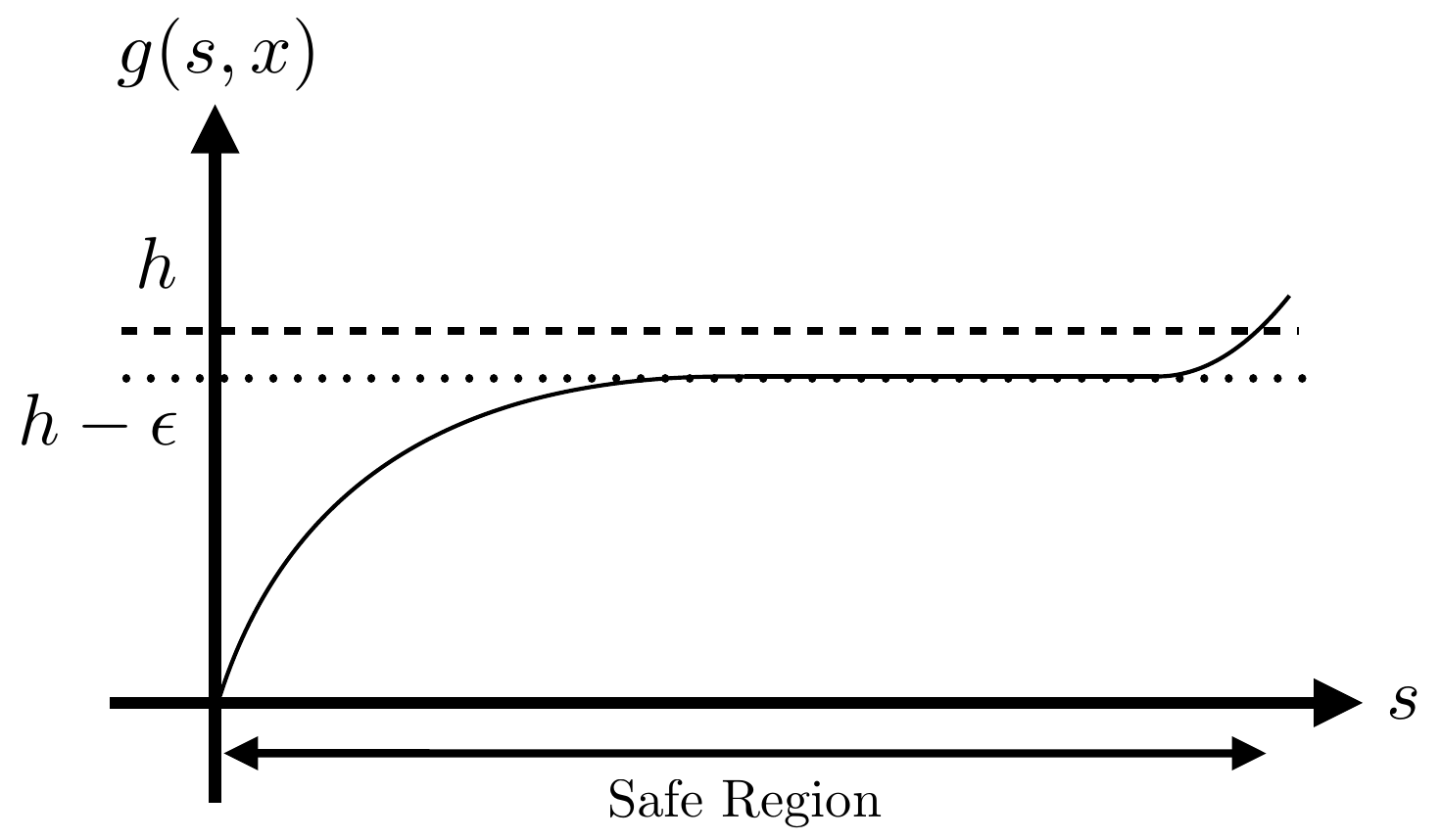}
        \par 
    \end{centering}
    
    \caption{Example of a function where identifying the safe boundary (i.e., the highest safe $s$) can be arbitrarily hard. \label{fig:necessity}}
\end{figure}

To see this, we consider the highly simplified special case in which there is only a \emph{single} choice of $\bx$ (i.e., $|\mathcal{D_X}|=1$), and the goal is to maximize $f$ with respect to $s \in [0,1]$ alone, subject to safety.  When $L'_g = 0$, we can encounter a situation such as that shown on Figure \ref{fig:necessity}, where $g$ is flat with a value extremely close to $h$, say $h-\epsilon$.  For arbitrarily small $\epsilon$, it is arbitrarily hard to identify (to within a constant accuracy, say $\pm 0.01$) the value of $s$ at which the function crosses from safe to unsafe.\footnote{For instance, supposing additive Gaussian noise, it is well-known that $\Theta\big( \frac{1}{\epsilon^2} \big)$ queries are needed to distinguish between function values of $h+\epsilon$ and $h-\epsilon$, and similarly for other values in between the two.}  On the other hand, if $f$ is \emph{not} flat (i.e, $L_f > 0$), then accurately identifying that crossing point is crucial for optimizing $f$.

Along similar lines, even if we have $L'_g > 0$, having $L_f = \infty$ would imply that even a minuscule amount of inaccuracy with respect to identifying the safe boundary (which is almost always unavoidable, particularly in the noisy setting) can lead to strict suboptimality with respect to $f$ due to abrupt changes.

Thus, the assumptions $L'_g > 0$ and $L_f < \infty$, or possibly similar kinds of assumptions with different specific details, are essential for the goals of our paper.  We note that while the assumption $L'_g > 0$ is somewhat specific to our setting, growth rate upper bounds (i.e., Lipschitz constants) are much more common, e.g., as used by existing algorithms such as \safeoptns.  Moreover, for many commonly-considered kernels (e.g., Mat\'ern with $\nu > 1$), Lipschitz continuity with respect to $s$ \emph{and} $\bx$ is automatically guaranteed for any function in the RKHS (e.g., see Remark 5 of \citet{shekhar2020multi}).

As a side note, we point out that strict monotonicity of $g(\cdot, \bx)$ alone may not directly imply $L_g' > 0$, either due to the presence of a global minimum at $g(0,\bx)$, or due to inflection points.  For example, $g(s,\bx) = s^2$ is strictly monotonically increasing in $s$, but the minimum value of the partial derivative with respect to $s$ is $0$.  While strict monotonicity still implies \eqref{eqn:lg'} for some $L'_g > 0$ when $s > s'$, the difference is that $L'_g$ needs to vary with $s'$ (and possibly approach 0 as $s'$ approaches $s$).

\subsection{Variations and Extensions} \label{app:extensions}

In our paper, we have modeled $f$ and $g$ separately for clarity of exposition.  However, we can easily handle joint modeling (e.g., to capture correlations between $f$ and $g$) in the same way as existing works such as \citep{berkenkamp2021bayesian}:  We simply define $h(\cdot,\cdot,1) = f(\cdot,\cdot)$ and $h(\cdot,\cdot,2) = g(\cdot,\cdot)$, and assume that the ``expanded'' function $h(s,\bx,i)$ has a low RKHS norm (instead of $f$ and $g$ separately).  The existing confidence bounds can then simply be applied to $h$, rather than to $f$ and $g$ separately.

In certain applications, the current problem setup may need to be extended to consider context variables $c_t$ that cannot be ``selected'', but are rather provided by the environment in every round (e.g., a patient's blood sugar level in the adaptive clinical trial application). Furthermore, safety may be dictated by multiple safety functions $g_1, g_2, \dots, g_l$, instead of a single function $g$. In both these scenarios, we note that our algorithms can be readily extended following the ideas proposed in \citep{berkenkamp2021bayesian}, assuming each $g_i$ satisfies our monotonicity assumption.

\end{document}